\documentclass[runningheads]{llncs}

\usepackage{eccv}

\usepackage{eccvabbrv}

\usepackage{graphicx}
\usepackage{booktabs}

\usepackage{float}
\usepackage{lipsum}
\usepackage{comment}
\usepackage{courier}
\usepackage{stackengine}%
\usepackage{amsmath}
\usepackage{lipsum} 
\usepackage{kotex}
\usepackage{comment}
\usepackage{wrapfig}
\usepackage{siunitx}
\usepackage{caption}
\usepackage[verbose]{microtype}
\usepackage[accsupp]{axessibility}  

\usepackage{hyperref}

\usepackage{orcidlink}

\begin{document}

\title{FlexiEdit: Frequency-Aware Latent Refinement for Enhanced Non-Rigid Editing} 

\titlerunning{FlexiEdit}

\author{Gwanhyeong Koo\orcidlink{0009-0005-6455-3223} \and
Sunjae Yoon\orcidlink{0000-0001-7458-5273} \and
Ji Woo Hong\orcidlink{0000-0002-3758-0307} \and 
Chang D. Yoo\orcidlink{0000-0002-0756-7179}}

\authorrunning{G. Koo et al.}

\institute{Korea Advanced Institute of Science and Technology, Daejeon, Republic of Korea
\email{\{kookie,sunjae.yoon,jiwoohong93,cd\_yoo\}@kaist.ac.kr}
}

\maketitle

\vskip -0.25in
\begin{figure*}[htb]
\begin{center}
\centerline{\includegraphics[width=0.9\textwidth]{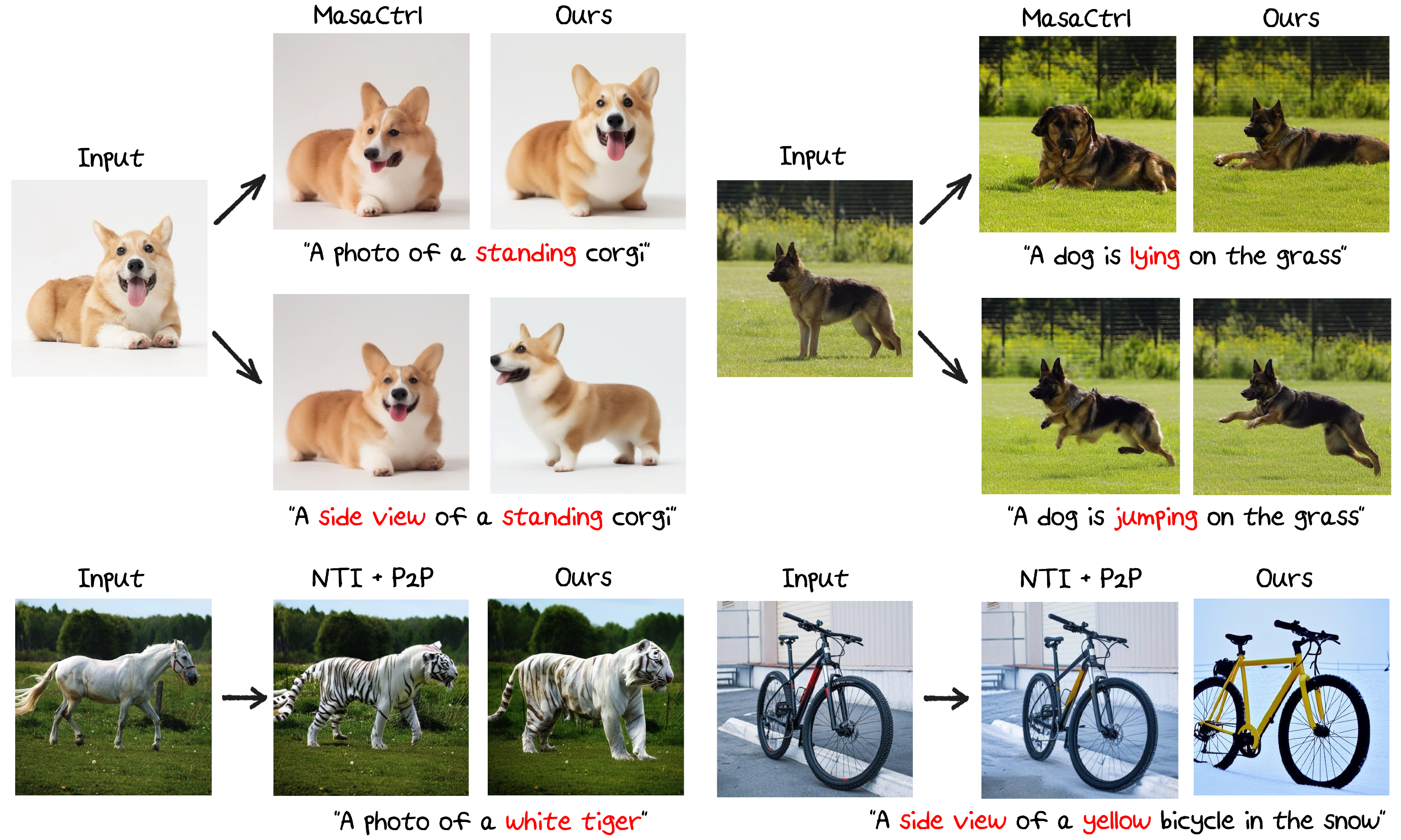}}
\caption{Comparative editing results using FlexiEdit (ours), MasaCtrl \cite{masactrl}, and Prompt-to-Prompt (P2P) \cite{prompt_to_prompt}. FlexiEdit outperforms other methods in non-rigid edits by providing more flexibility in altering layouts and achieving more natural results in rigid edits.}
\label{fig:0_teaser}
\end{center}
\vskip -0.5in
\end{figure*}

\begin{abstract}
  Current image editing methods primarily utilize DDIM Inversion, employing a two-branch diffusion approach to preserve the attributes and layout of the original image. However, these methods encounter challenges with non-rigid edits, which involve altering the image's layout or structure. Our comprehensive analysis reveals that the high-frequency components of DDIM latent, crucial for retaining the original image's key features and layout, significantly contribute to these limitations. Addressing this, we introduce \textbf{FlexiEdit}, which enhances fidelity to input text prompts by refining DDIM latent, by reducing high-frequency components in targeted editing areas. FlexiEdit comprises two key components: (1) Latent Refinement, which modifies DDIM latent to better accommodate layout adjustments, and (2) Edit Fidelity Enhancement via Re-inversion, aimed at ensuring the edits more accurately reflect the input text prompts. Our approach represents notable progress in image editing, particularly in performing complex non-rigid edits, showcasing its enhanced capability through comparative experiments.
\end{abstract}

\begin{figure}[t!]
\centering
    \includegraphics[width=1.0\textwidth]{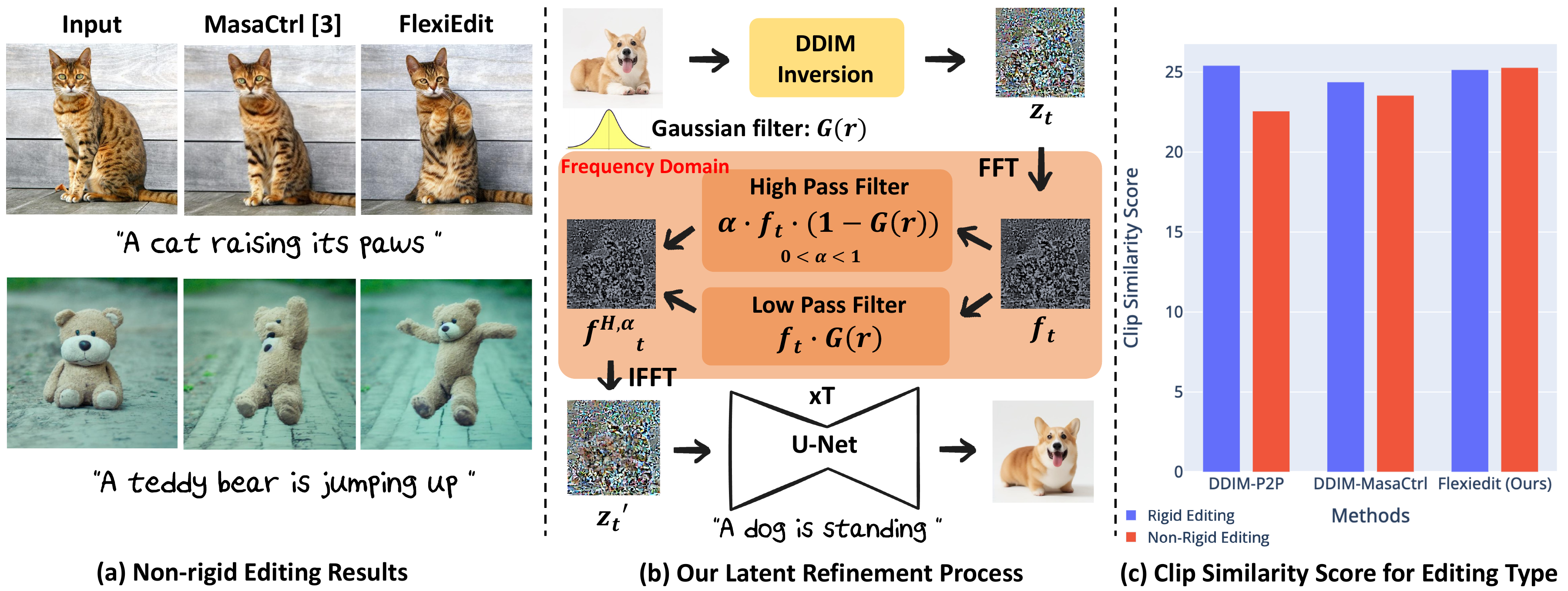}
   \caption{(a) Comparison of non-rigid edit outcomes between MasaCtrl \cite{masactrl} and FlexiEdit, showing FlexiEdit's enhanced flexibility. (b) A schematic of Latent Refinement in FlexiEdit, illustrating the reduction of high-frequency components in the original latent for improved non-rigid editing. (c) Comparative CLIP similarity scores for P2P \cite{prompt_to_prompt}, MasaCtrl \cite{masactrl}, and FlexiEdit in rigid and non-rigid edits on the PIE benchmark \cite{proxedit}.}
\label{fig:1_introduction}
\end{figure}

\section{Introduction}
\label{sec:intro}

Diffusion models \cite{ddpm} have achieved significant progress beyond Generative Adversarial Networks (GANs) \cite{gan_2014, stackgan, gan_overview, gan_2019} in the domain of Text-to-Image (T2I) generation. Models trained on extensive datasets \cite{dalle2, imagen, glide, stylediffusion, ldm}, notably Stable Diffusion \cite{ldm}, have been widely recognized for their ability to generate high-quality images from text descriptions. This notable success of these T2I models has naturally led to an extension of research towards image editing. As a technology that enables users to modify existing original images according to their preferences, image editing has become an important tool in our daily interactions with visual content. However, it has emerged that existing image editing methods encounter limitations in performing flexible editing tasks, such as non-rigid edits (e.g., pose, view change). 

Current research in image editing primarily utilizes DDIM Inversion \cite{ddim} for editing while preserving the original image. This approach ensures the edited image retains the original's attributes and layout by injecting attention features \cite{prompt_to_prompt, plug_and_play, masactrl, fatezero}. Alongside, inversion methods \cite{nti, npi, direct_inversion, proxedit} that aim to closely apply the original image to the editing target have been extensively explored. These inversion techniques, combined with editing methods, have demonstrated excellent results in rigid edits aimed at preserving the original image's structure. However, while these editing and inversion approaches achieve high fidelity to the original image, they struggle with non-rigid edits, such as changing the image's layout. To address the limitations in non-rigid editing, methods involving fine-tuning and the precise injection of attention features have been introduced. Imagic \cite{imagic}  requires fine-tuning the entire model and optimizing textual embedding for each input image, which can demand significant resources. On the other hand, MasaCtrl \cite{masactrl} enables non-rigid edits without fine-tuning, though it may result in minimal alterations to the original layout or fail in more flexible pose or motion change.


In this study, we discover that existing image editing methods struggle with non-rigid edits due to the DDIM latent space retaining the original image's attributes and layout, motivated by findings in \cite{freeinit}. Our exploration into the frequency components of the DDIM latent revealed that its high-frequency elements contain essential information about the layout. This observation indicates that the high-frequency components in the DDIM latent hinder flexible editing. Building on these findings, we introduce \textbf{FlexiEdit}, a novel image editing approach that refines DDIM latent to surpass these limitations, significantly enhancing layout editing flexibility while preserving key attributes. FlexiEdit consists of the following two key features: (1) Latent Refinement: It reduces the high-frequency components and adds Gaussian noise within the DDIM latent designated for editing region, as illustrated in Fig \ref{fig:1_introduction} (b), enabling the formation of layouts different from the original image. (2) Edit Fidelity Enhancement via Re-inversion: This process enhances the two-diffusion branches by focusing on two main goals. Firstly, it aims to maximize the effectiveness of edits within the target branch. Secondly, it ensures the preservation of the original object's attributes through a novel re-inversion process. This dual approach intensifies the editing capabilities in the target branch without initially relying on direct feature injection from the source branch. After the image is generated in the target branch, it undergoes re-inversion. Subsequently, in the resampling phase, features from the source branch are seamlessly injected, infusing the attributes of the original image.

In comparative experiments with other image editing methods, FlexiEdit has demonstrated outstanding performance, particularly in non-rigid edits. It has also excelled in preserving content and maintaining fidelity during editing, as evidenced by evaluations on the PIE bench dataset \cite{direct_inversion}.

\section{Related Works}

\subsection{Text-guided Image Editing}
In the field of text-guided image editing, initial approaches often relied on additional masks with text inputs to edit specific parts of the image \cite{glide, blended_diffusion}. Subsequent research introduced DDIM Inversion, allowing unedited regions to remain unchanged without the need for masks or additional guidance \cite{diffedit, diffusion_clip}. At the same time, the success of large-scale text-to-image (T2I) generation models \cite{dalle2, ldm, imagen, glide}, such as stable diffusion \cite{ldm}, facilitated the utilization of pretrained T2I models in image editing. This gave rise to the two-diffusion branch methodology, where the source branch reconstructs the original image while the target branch generates the edited image. Within this framework, research has been conducted to inject either (1) cross-attention maps \cite{prompt_to_prompt, masactrl, fatezero} or (2) spatial features from residual and self-attention blocks \cite{plug_and_play} from the source to the target branch. Based on these T2I image editing methods, various techniques have extended to video editing \cite{tune_a_video, tokenflow, frag}.

However, previous methods faced challenges in achieving flexible editing tasks like non-rigid edits, as they directly applied the features of the original image to the edited image. Imagic \cite{imagic} addressed this by fine-tuning the entire model and optimizing textual embeddings for each input image to perform non-rigid edits. Meanwhile, MasaCtrl \cite{masactrl} modified the self-attention mechanism in the target branch to perform non-rigid edits without fine-tuning. However, it often encountered limitations and failures when objects within the image underwent significant changes.

\subsection{Inversion methods in Image Editing}
In using DDIM Inversion \cite{ddim} for image editing, there exists a drawback: it cannot entirely reconstruct the original image when the classifier-free guidance (CFG) \cite{cfg} scale is greater than 1. DDIM Inversion assumes an ODE (Ordinary Differential Equation) reversal within very small steps during the DDIM sampling process, resulting in an approximation of the solution to the Neural ODE via Euler's method. Due to this approximation in the ODE, a slight error accumulates during denoising. Moreover, while DDIM Inversion adds noise to the original image with a CFG scale of 1, the DDIM sampling process operates with a CFG scale greater than 1 \cite{diffusion_beat_gans, cfg} to apply edits different from the original image. This disparity contributes to the accumulation of errors compared to the latent trajectory obtained through DDIM Inversion. Consequently, the reconstruction from the source branch is not well-performed, leading to suboptimal editing performance in the target branch.

To address this, efforts have been made to align the DDIM Inversion trajectory with the DDIM sampling trajectory to enable a complete reconstruction of the original image when CFG scale exceeds 1. NTI \cite{nti} proposed an optimization-based inversion method that optimizes null text used in classifier-free guidance. However, due to the time-consuming nature of the optimization process, research has been conducted to achieve similar effects while finding optimal timesteps \cite{object_aware_inversion, wavelet}. Additionally, approaches have been presented to recover the original image without optimization \cite{npi, proxedit, direct_inversion}. These inversion methods can be integrated with image editing methods to enhance their capabilities.

\section{Preliminaries and Observations}
\subsection{DDIM Inversion}

DDIM extends DDPM into a non-Markovian diffusion process, enabling the training of a deterministic generative process. Within the framework of LDM, deterministic DDIM sampling employs a denoiser network ${\epsilon_\theta}$, denoted as follows:

\begin{equation}
\begin{aligned}
z_{t-1} = \sqrt{\frac{\alpha_{t-1}}{\alpha_{t}}}z_{t} + \sqrt{\alpha_{t-1}} \left( \sqrt{\frac{1}{\alpha_{t-1}}-1} - \sqrt{\frac{1}{\alpha_{t}}-1} \right) \epsilon_\theta(z_t, t), \label{eq:1}
\end{aligned}
\end{equation}

where ${\epsilon_\theta}$, is utilized to predict $\epsilon(z_t, t)$ at each timestep, ranging from 1 to $T$. Here, $z_t$ represents the latent variable at timestep $t$. This approach facilitates image generation from random Gaussian noise $z_T$. By rephrasing the DDIM sampling equation within an ordinary differential equation (ODE), Euler Integration can be applied to solve the ODE for the reverse process. This adaptation allows the encoding from $z_0$ to $z_T$, referred to as DDIM Inversion:

\begin{equation}
\begin{aligned}
{z^*_t} = \sqrt{\frac{\alpha_{t}}{\alpha_{t-1}}}{z^*_{t-1}} + \sqrt{\alpha_{t}} \left( \sqrt{\frac{1}{\alpha_{t}} - 1} - \sqrt{\frac{1}{\alpha_{t-1}} - 1} \right) \epsilon_\theta({z^*_{t-1}}, t). \label{eq:2}
\end{aligned}
\end{equation}

In Eq \ref{eq:2}, $z^*_t$ denotes latent features during the DDIM Inversion process. Therefore, in the process of inverting the original image, we obtain the DDIM Inversion trajectory, denoted as $[z^*_t]^{T}_{t=0} $. 
Following this, by initiating the DDIM sampling from ${{z}_T} = {z^*_T}$, a reconstruction trajectory of $[z_t]^{t=T}_{0} $ is achieved. As CFG scale is greater than 1, errors accumulate during this process. Consequently, the disparity between ${z^*_t}$ and ${{z}_t}$ gradually increases as the denoising progresses.

\subsection{Frequency Analysis of DDIM Latent: : Unveiling the Role of High Frequencies}
\label{sect:3.2}

In this section, a frequency analysis is conducted to investigate which components of the DDIM latent $z_T$ contribute to preserving the attributes and layout of the original image during the image reconstruction process. Our methodology begins by separating the DDIM latents into high and low frequency components within the frequency domain. Let the original image be denoted by $I_{src}$, and its encoded latent by $z_0$. The derivation of the DDIM latent $z_T$ from $z_0$ is achieved through the DDIM Inversion process, as detailed in section 3.1 (Eq \ref{eq:2}), and the process of transforming $z_T$ into the frequency domain is achieved using Fourier transform (Eq \ref{eq:3}). Here, $FFT(·)$ and $IFFT(·)$ correspond to the 2D Fast Fourier Transform, and its inverse, $f_T$ is the frequency domain counterpart of $z_T$.

\begin{align}
& f_T = FFT(z_T) \label{eq:3}, \\
& \mathcal{L}_r = \frac{1}{2\pi\sigma^2}e^{-\frac{r^2}{2\sigma^{2}}} \in \mathbb{R}^{W \times H}, \quad  \mathcal{H}_r = 1 - \mathcal{L}_r  \in \mathbb{R}^{W \times H}, \label{eq:4}
\end{align}

To analyze frequency components of $z_T$, we apply a 2-dimensional Gaussian low-pass filter $\mathcal{L}_r \in \mathbb{R}^{W \times H}$ and a high pass filter $\mathcal{H}_r \in \mathbb{R}^{W \times H}$, where W and H represent the filter’s width and height. Additionally, $r= \sqrt{x^2 + y^2}$ denotes the distance from the center of the Gaussian filter to each point (x, y), with $\sigma$ acting as a scaling coefficient for the Gaussian curve. Utilizing $\mathcal{L}_r$ and $\mathcal{H}_r$, the low and high-frequency components of $f_T$ are separated. A scalar $\alpha$ ranging from 0 to 1, is applied to both components, but it modifies only the low-frequency component in one instance and solely the high-frequency component in another. This method results in $f^{L, \alpha}_{T}$ for the low-frequency adjustments and $f^{H, \alpha}_{T}$ for the high-frequency modifications (Eq \ref{eq:5}, \ref{eq:6}). Here, $\odot$ denotes the element-wise multiplication operation, used to apply the low-pass and high-pass filter to $f_T$.

\begin{figure}[t!]
\centering
    \includegraphics[width=0.9\textwidth]{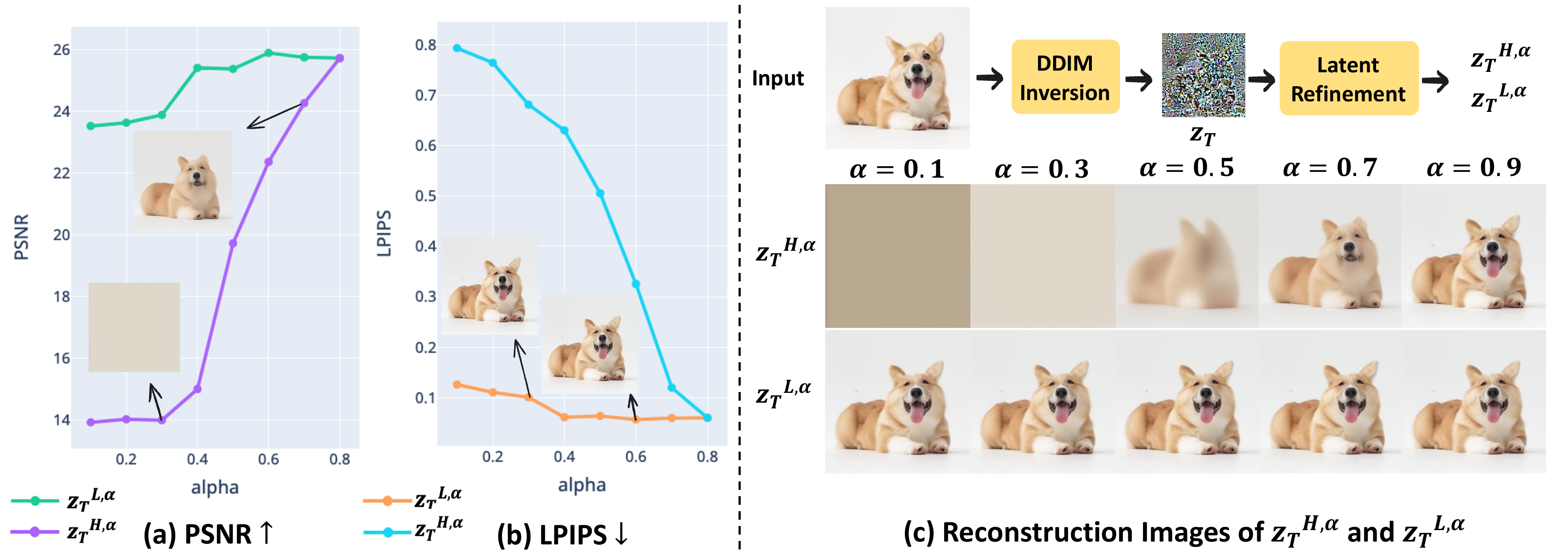}
   \caption{(a), (b) Show the PSNR and LPIPS results of reconstructing $z^{H, \alpha}_T$, and $z^{L, \alpha}_T$ in comparison to the original image. (c) visualizes the reconstruction outcome across different alpha values, indicating that high-frequency components play a more significant role in forming the object's layout than low-frequency components.}
\label{fig:3_preliminary}
\end{figure}




\begin{align}
& f^{L, \alpha}_T = \alpha * f_T \odot \mathcal{L}_r + f_T \odot \mathcal{H}_r  \quad where \quad \alpha \: \in \: [0, 1],  \label{eq:5} \\
& f^{H, \alpha}_T = f_T \odot \mathcal{L}_r + \alpha * f_T \odot \mathcal{H}_r  \quad where \quad \alpha \: \in \: [0, 1], \label{eq:6} \\ 
& z^{H, \alpha}_T = IFFT(f^{H, \alpha}_T), \quad z^{L, \alpha}_T = IFFT(f^{L, \alpha}_T). \label{eq:7}
\end{align}

The resultant $z^{H, \alpha}_T$ represents a latent with reduced high-frequency components compared to the original $z_T$, while $z^{L, \alpha}_T$ indicates a latent with less low-frequency components (Eq \ref{eq:7}). Adjusting the scalar $\alpha$ to modulate frequency component reduction, reconstructions were carried out from $z^{H, \alpha}_T$, and $z^{L, \alpha}_T$. Subsequent to this, an evaluation of PSNR and LPIPS \cite{psnr_lpips} for the reconstructed images against the original was conducted.  In Fig \ref{fig:3_preliminary} (a) and (b), we observe that $\alpha$ increases in reconstructions from $z^{H, \alpha}_T$, there is a notable improvement in image quality, as demonstrated by higher PSNR and lower LPIPS values. In contrast, reconstructions from $z^{L, \alpha}_T$ not only exhibit slight variations in PSNR and LPIPS values across different $\alpha$ levels but also resemble the original image in visual appearance. This distinction indicates that high-frequency elements within $z_T$ are more crucial in determining the attributes and layout of the original image than low-frequency elements.

\section{Method}
Given the findings from Section \ref{sect:3.2}, which observed the high-frequency component of $z_T$ as imposing the attributes and layout of the original image, it becomes evident why DDIM Inversion-based image editing methods face challenges with non-rigid editing. The persistence of the original image's elements within $z_T$ presents a significant obstacle. Motivated by these insights, we introduce FlexiEdit, a method designed to enhance the flexibility of non-rigid edits. FlexiEdit is comprised of two strategies: (1) Latent Refinement and (2) Edit Fidelity Enhancement via Re-inversion.

\begin{figure}[t!]
\centering
    \includegraphics[width=1.0\textwidth]{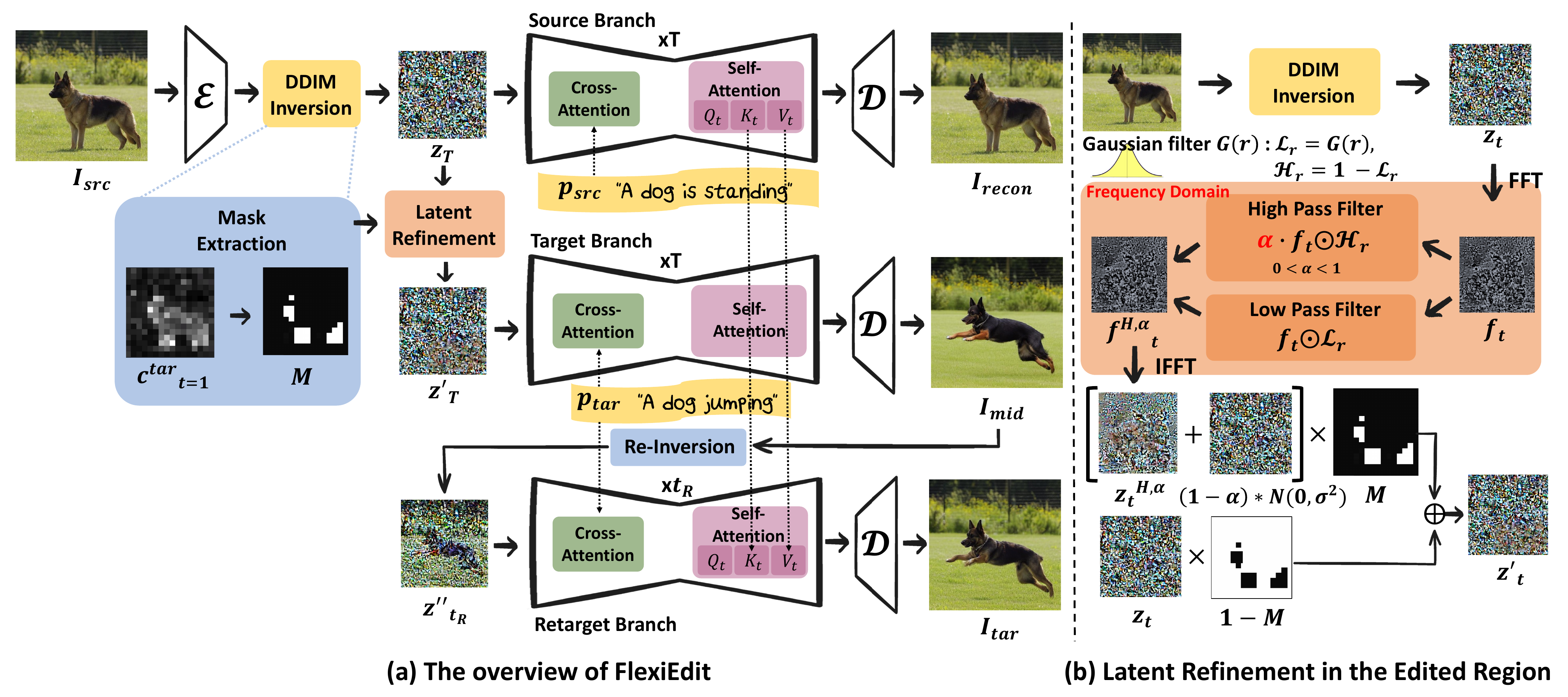}
   \caption{\textbf{The pipeline of FlexiEdit}. (a) Our method utilizes the refined latent $z'_{T}$ to achieve $I_{mid}$, which significantly alters the original image’s layout. Following re-inversion over a duration of $t_R$, features from the original image are injected during the resampling process, resulting in the final edited image, $I_{tar}$. (b) The refinement process within the edited region of the latent entails reducing high-frequency components by a factor of $\alpha$ while incorporating Gaussian noise proportional to $(1 - \alpha)$.}
\label{fig:4_model}
\end{figure}

\subsection{Latent Refinement} 
\label{sect:4.1}
In image editing, it's essential to preserve the integrity of unedited regions. Thus, we implement Latent Refinement in the designated editing areas, further incorporating Gaussian noise to facilitate more natural changes in the object's layout. Consider an input image $I_{src}$ with source prompt $p_{src}$ and a target prompt $p_{tar}$, aiming for an edit towards $I_{tar}$. The editing region is specified as binary mask $M$, established via two approaches: (1) leveraging cross-attention maps and (2) incorporating user input. To obtain the $M$ from the cross-attention maps for the edited words, the initial step involves distinguishing the edited word by comparing $p_{src}$ and $p_{tar}$. We selected words present in $p_{tar}$ but absent in $p_{src}$ and then measured the CLIP similarity score between $I_{src}$ and these words. Words with a similarity score below a certain threshold are designated as the edited words $w_{ed}$. During the DDIM Inversion process, when the $I_{src}$ is transformed into $z_T$, both $p_{src}$ and $p_{tar}$ are applied (Eq \ref{eq:8}). For generating the mask $M$ for the edited words, we calculate the average of their cross-attention maps $[c^{w_{ed}}_t]^T_{t=1} \in \mathbb{R}^{16 \times 16 \times N}$ at a 16x16 spatial resolution across all UNet layers, where $N$ represents the number of tokens in $p_{tar}$. Our experiments confirm that employing just $[c^{w_{ed}}_t]_{t=1}$ is adequate for precisely capturing the attention mask that correlates with the edited words. Subsequently, a predetermined threshold is applied to these averaged maps, converting them into a binary format to finalize the mask $M$ for the edited words (Eq \ref{eq:9}). However, this method is based on the CLIP similarity score, making it dependent on the threshold value. Additionally, there are instances where the cross-attention map for the edited words did not correspond to the area we actually wanted to edit. Therefore, we also utilize an approach allowing users to directly select the region to be edited on the original image to obtain the mask.

\begin{figure}[t!]
\centering
    \includegraphics[width=0.9\textwidth]{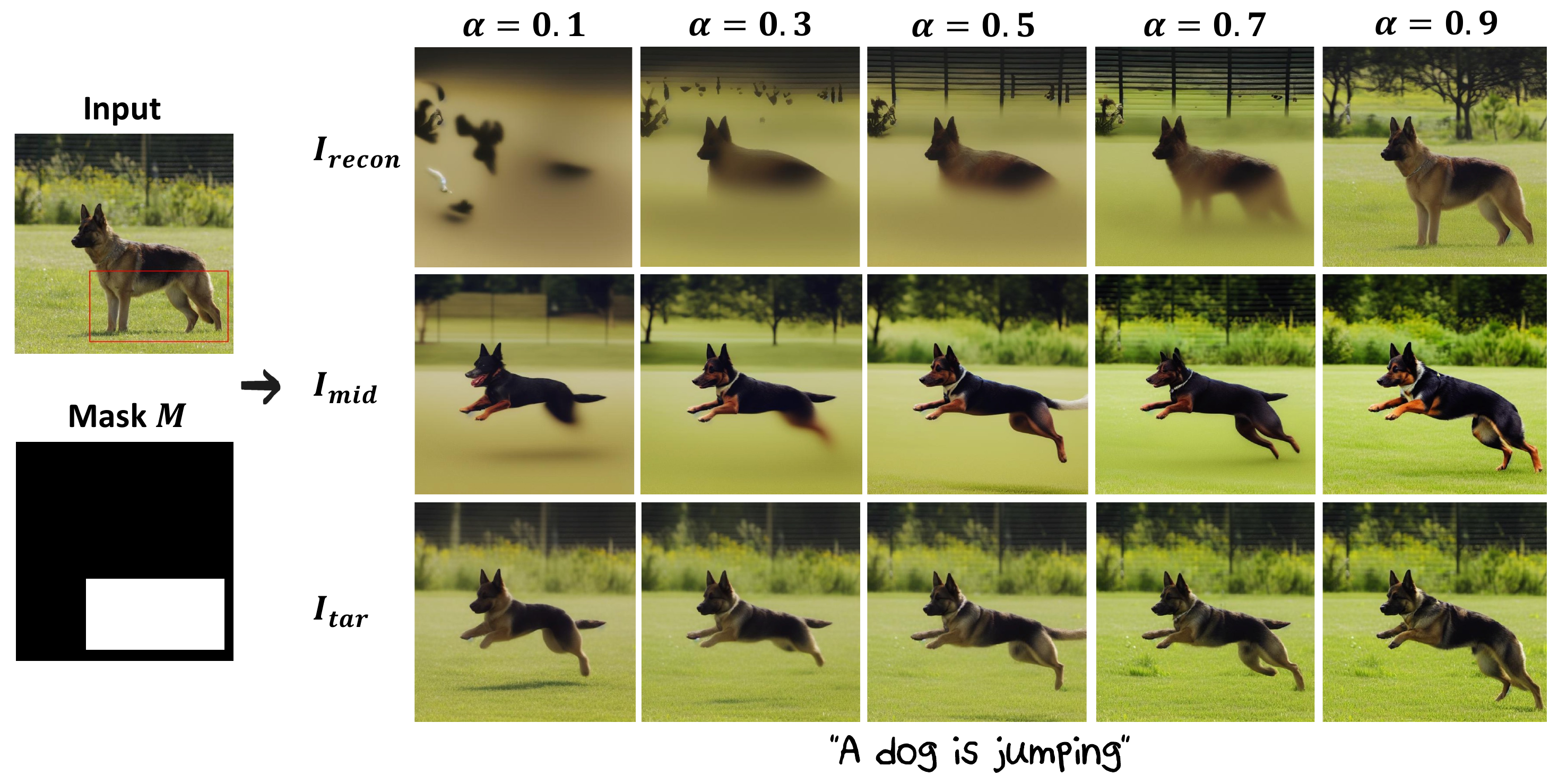}
   \caption{Illustrates the results of adjusting $\alpha$ values on latent refined within the user mask $M$ region, resulting in $I_{recon}$, $I_{mid}$, and $I_{tar}$.  As the $\alpha$ value decreases, there are more significant deviations from the original image's layout. In contrast, higher $\alpha$ values result in a layout that closely aligns with the original image.}
\label{fig:5_experiment_alpha}
\end{figure}

\begin{align}
& z_{T}, [c^{p_{tar}}_t]^T_{t=1} = \text{DDIM-Inv}(z_{0}, p_{src}, p_{tar}), \label{eq:8} \\
& M = \text{Mask-Extraction} ([c^{w_{ed}}_{t}]_{t=1}) \in \mathbb{R}^{16 \times 16 \times 1}, \label{eq:9} \\
& z'_T = z_T * (1-M) + (z^{H, \alpha}_T + \mathcal{N}(0, \sigma^2)* (1-\alpha)) * M \label{eq:10}.
\end{align}

After acquiring the mask $M$, we utilize $z^{H, \alpha}_T$ blended with Gaussian noise scaled by ${1-\alpha}$ in the target editing area $M$, while retaining the $z_T$ in the $1-M$ (Eq \ref{eq:10}). This procedure selectively reduces the frequency components by a factor of $\alpha$ and introduces Gaussian noise scaled by ${1-\alpha}$ exclusively in the region to be edited, resulting in a refined latent representation, $z'_T$. By employing $z'_T$ for image editing, we facilitate the flexible modification of the object layout within the edit region, allowing for adaptive and seamless editing tailored to the specific editing objectives. When the user defines the mask $M$ over the desired region, the variation according to the value of $\alpha$ is illustrated in Fig \ref{fig:5_experiment_alpha}.

\subsection{Edit Fidelity Enhancement via Re-inversion}
In advancing beyond the two-branch diffusion methods utilized in the current image editing methodologies, we introduce a novel three-branch approach leveraging Re-inversion, which consists of the following components. Illustrated in Fig \ref{fig:4_model}, the source branch reconstructs $I_{recon}$. The target branch is designated for producing $I_{mid}$, and the retarget branch focuses on generating the ultimate desired image, $I_{tar}$. 

\paragraph{\bf Source and Target Branch.} 
The source branch begins with inputs $z_T$ and $p_{src}$, engaging in reconstruction to accurately restore the original image, achieved by setting the CFG scale to 1. In contrast, the target branch processes input $z'_T$ and $p_{tar}$ to perform editing. Here, the CFG scale is set to 7.5, and it operates independently of the key and value features from the source branch. This independence ensures that edits deviate maximally from the original image. The resultant image, $I_{mid}$, exhibits altered layouts, diverging slightly from the original. Subsequently, this image undergoes Re-inversion over a duration $t_R$.

\paragraph{\bf Re-inversion and Retarget branch.} 
The re-inverted latent $z''_{t_R}$ (Eq \ref{eq:11}) is then processed through the retarget branch, where it is denoised in a UNet with $p_{tar}$. Importantly, this stage incorporates key and value feature injections from the source branch, integrating characteristics of the original image (Eq \ref{eq:12}). The Re-inversion process is formalized as follows.

\begin{align}
& z''_{t_R} = \text{DDIM-ReInv}(z'_{0}, p_{tar}, t_{R}), \label{eq:11} \\
& \text{Attention}(Q, K, V) = \text{Softmax}\left(\frac{QK_{\text{src}}}{\sqrt{d}}\right) \cdot V_{\text{src}}.  \label{eq:12}
\end{align}

The objective of the Retarget branch is not to edit but to maximize the retention of features from the original image, with feature injections applied throughout the denoising steps. Determining the optimal $t_R$ is crucial in this context, as it directly influences how the original image's features are preserved and integrated into the final edited image. The choice of $t_R$ within the range $[1, T]$ depends on the size of the edit region and the degree of similarity between $I_{mid}$ and $I_{src}$. A longer $t_R$ is required for larger edit regions or when $I_{mid}$ substantially differs from $I_{src}$, to ensure thorough integration of original features. In contrast, smaller edit regions that are more closely aligned with the original necessitate a shorter $t_R$. Based on many experimental results with various values of $t_R$, we discover that the optimal $t_R$ value varies depending on the type of image editing, and we have set the range of $t_R$ to $[t_{R1}, t_{R2}]$. Detailed explanations are included in the ablation study, section \ref{ablation:t_R}. Therefore, optimal $t_R$ ranges between $t_{R1}$ and $t_{R2}$, with the duration finely adjusted according to both the size of the edit region and the similarity between $I_{mid}$ and $I_{src}$. To precisely adjust $t_R$, we consider the edit region’s proportion by using the total area of mask $M$, denoted as $A_{total}$, and the area within $M$  denoted as $A_{edit}$. This approach enables the calibration of $t_R$ in relation to the edited area’s size. Furthermore, to assess the degree of similarity between the original and edited images, we examine the ratio of PSNR values, \(\text{PSNR}(I_{\text{src}}, I_{\text{recon}})\) and \(\text{PSNR}(I_{\text{src}}, I_{\text{mid}})\). This comparison aids in evaluating how significantly $I_{mid}$ has altered from $I_{src}$, reflecting the impact of the editing process. We have set the coefficients $\alpha_R$ and $\beta_R$ to 0.5 each, ensuring a balanced consideration of the edit region's size and the similarity in determining $t_R$ (Eq \ref{eq:13}). Using the determined $t_R$, the latent $z''_{t_R}$ undergoes denoising for a duration of $t_R$ steps, resulting in the final results, $I_{tar}$.

{\scriptsize
\begin{align}
& t_R = t_{R1} + (t_{R1} - t_{R2}) \cdot \left( \alpha_R \cdot \left( \frac{A_{\mathrm{edit}}}{A_{\mathrm{total}}} \right) + \beta_R \cdot \left( 1 - \frac{\mathrm{PSNR}(I_{\mathrm{src}}, I_{\mathrm{mid}})}{\mathrm{PSNR}(I_{\mathrm{src}}, I_{\mathrm{recon}})} \right) \right). \label{eq:13}
\end{align}
}

\section{Experiments}
\subsection{Implementation Details}
\paragraph{\bf Setup}
In the development of FlexiEdit, we employ the Latent Diffusion Model (LDM) \cite{ldm} leveraging the publicly available Stable Diffusion v1.4 checkpoint. For the sampling process, we utilize a DDIM schedule with $T=50$ steps. In terms of our FlexiEdit model, the source branch operates with a CFG scale set to 1. In contrast, we apply a CFG scale of 7.5 in both the target and retarget branches. Feature injection from the source branch to the retarget branch is carried out from the 0 to the $t_R$ denoising step within the UNet’s decoder. To address low-pass and high-pass filtering, we set the parameter value for $\sigma=0.3$, enabling adequate distinction between low and high-frequency components. Furthermore, the application of the $\alpha$ value on $z^{H, \alpha}_T$ significantly impacts the preservation of the original image's layout. As depicted in Fig \ref{fig:5_experiment_alpha}, a higher $\alpha$ value retains more of the original layout, whereas a lower $\alpha$ value induces more layout changes. We have designated this $\alpha$ as a hyperparameter, allowing users to adjust it according to their desired extent of layout modification. Through extensive experimentation, we determine that setting the $\alpha$ value within the range of $[0.5, 0.9]$ optimally preserves the original image's features while allowing for diverse layout changes. All experiments are conducted on NVIDIA A100 GPUs.

\paragraph{\bf Baselines and Dataset.}
For a detailed evaluation of FlexiEdit's performance, we compare it against current state-of-the-art (SOTA) image editing methods, such as Prompt-to-prompt (P2P) \cite{prompt_to_prompt} and plug-and-play (PnP) \cite{plug_and_play}, as well as methods capable of non-rigid editing like MasaCtrl \cite{masactrl} and ProxMasaCtrl \cite{proxedit}. The inversion methods for P2P and PnP utilize a direct inversion \cite{direct_inversion} approach, while MasaCtrl and ProxMasaCtrl are evaluated using a standard DDIM inversion \cite{ddim} method. For this analysis, we selected PIE-Bench as our dataset, providing a benchmark for a wide range of image editing tasks. Specifically, to assess non-rigid editing capabilities, we focus our experiments on a strategically selected subset of 30 images from PIE-Bench \cite{direct_inversion} and ELITE \cite{elite}, employing prompts designed explicitly for non-rigid edits.

\paragraph{\bf Evaluation Metrics.}
To compare the performance of different methods, we utilize six metrics. Structure Distance \cite{structure_distance} evaluates the structural similarities to the original images, focusing on structural aspects beyond appearance. For background preservation, we measure performance using PSNR, LPIPS \cite{psnr_lpips}, MSE, and SSIM \cite{ssim}. The text-image consistency is assessed using CLIP similarity \cite{clip}, where evaluations are conducted separately on the whole image and the editing mask to ensure a thorough analysis. Detailed descriptions of each metric are included in the supplementary file.

\begin{figure}[t!]
\centering
    \includegraphics[width=0.74\textwidth]{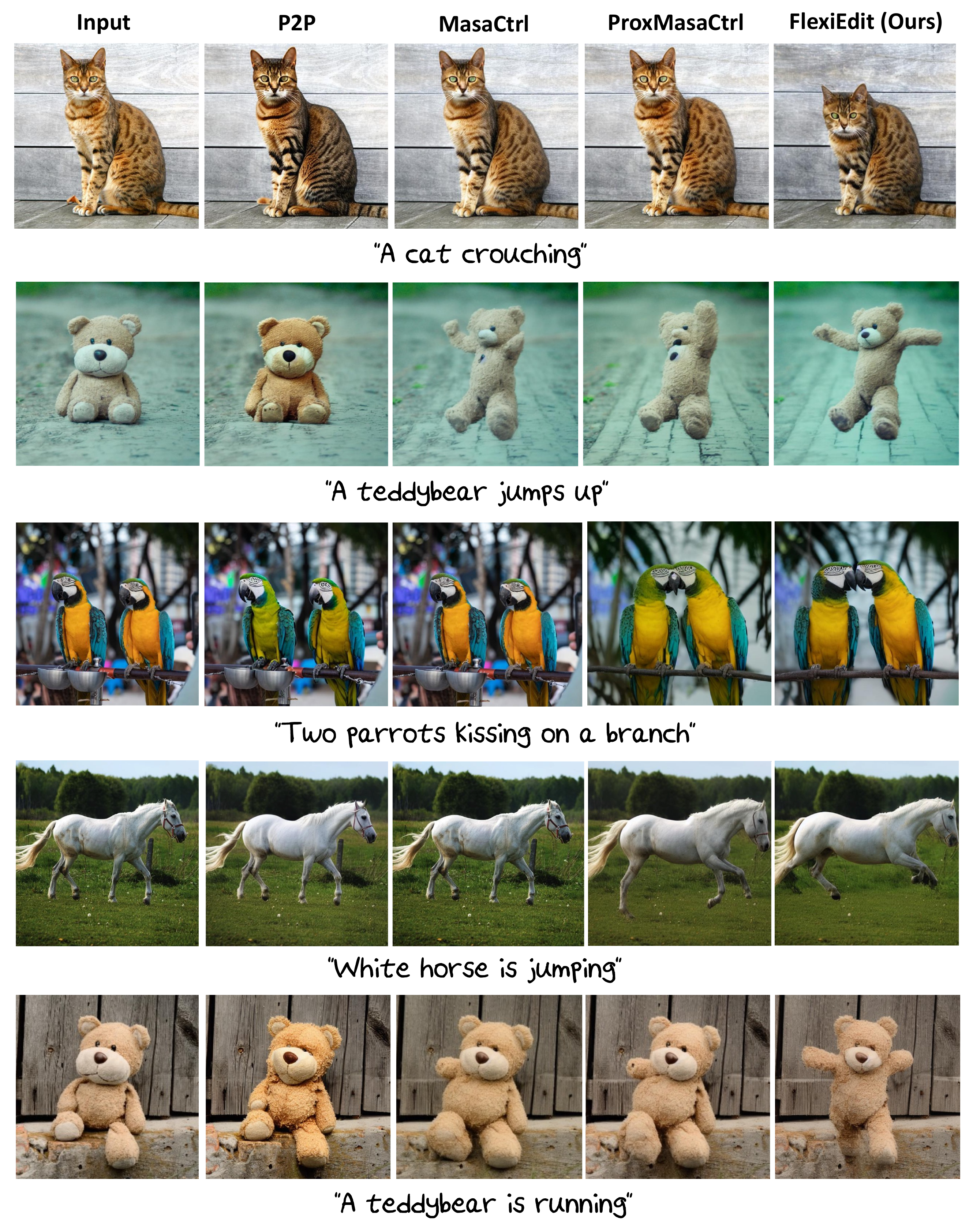}
   \caption{\textbf{Non-rigid Editing Results.} We compare the outcomes of non-rigid editing across current methods and FlexiEdit. P2P \cite{prompt_to_prompt} struggles to change the original layout, while MasaCtrl \cite{masactrl} and ProxMasaCtrl \cite{proxedit} make modifications that are awkward or slight. FlexiEdit excels at flexibly altering the layout to match the user's input text prompt.}
\label{fig:6_experiment_non_rigid_edit}
\end{figure}

\subsection{Comparisons with other image editing methods}
\paragraph{\bf Non-rigid Editing Results}

\begin{figure}[b!]
\centering
    \includegraphics[width=0.73\textwidth]{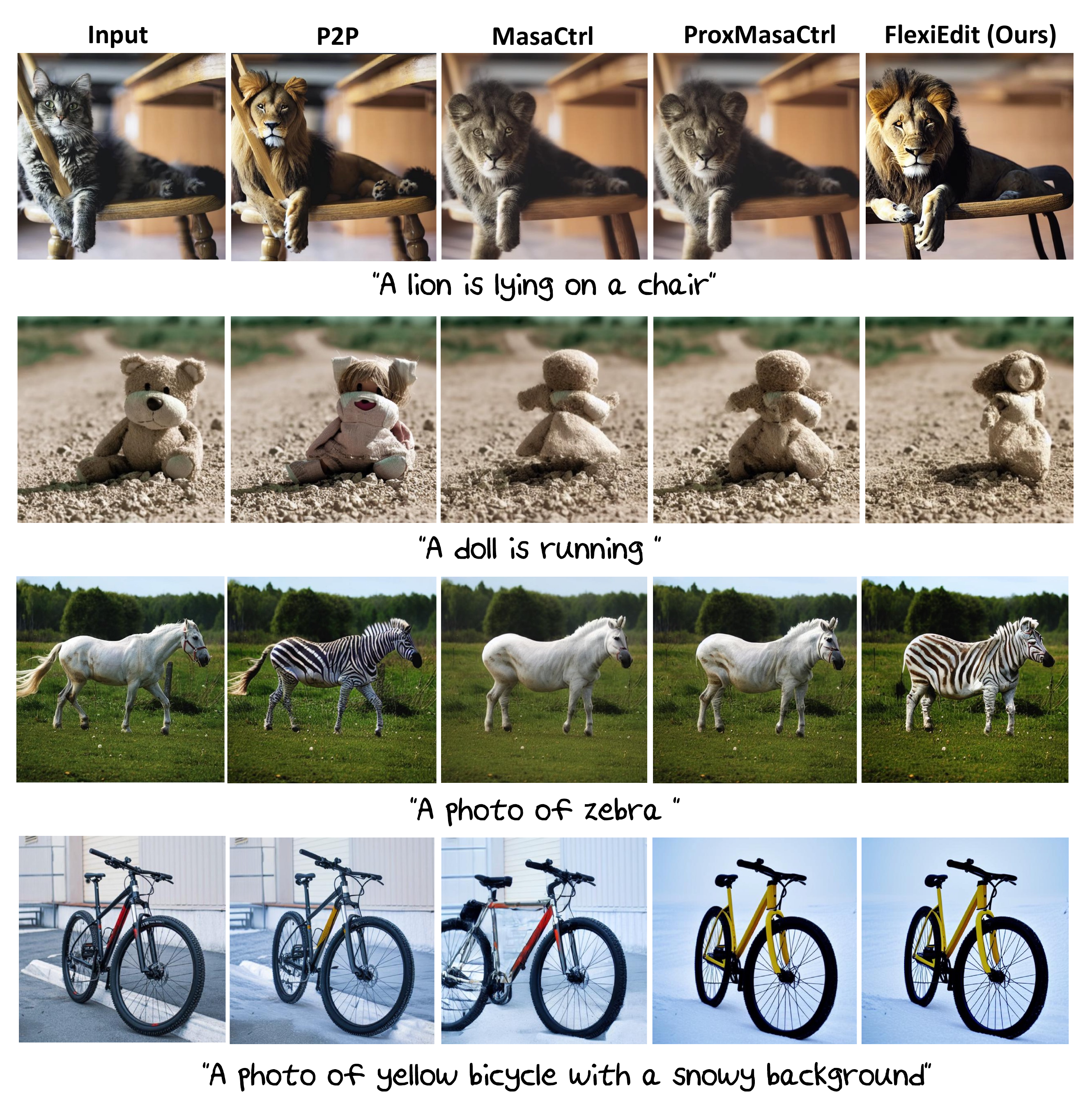}
   \caption{\textbf{Rigid Editing Results.} We assess rigid editing by comparing current methods and FlexiEdit. P2P \cite{prompt_to_prompt} edits without significantly deviating from the original layout, while MasaCtrl \cite{masactrl} and ProxMasaCtrl \cite{proxedit} fail to achieve object changes. In contrast, our method flexibly transforms the layout to align with the input text prompt, while preserving the original image's characteristics.
}
\label{fig:7_experiment_rigid_edit}
\end{figure}


In this section, we compare the non-rigid editing results between FlexiEdit and other image editing methods. All examples shown in Fig \ref{fig:6_experiment_non_rigid_edit} are pose changes, showing various instances of non-rigid edits. Notably, FlexiEdit excels in flexibly changing the layout while keeping the attributes of the image. In contrast, P2P struggles to change the layout of objects within the original image significantly. MasaCtrl and ProxMasaCtrl can adjust the object's layout, but these changes are either limited or result in awkwardness and artifacts. Our method shows superior performance in doing non-rigid edits, allowing for more flexible changes of object layouts from the image. The qualitative results, as seen in Table \ref{table:1}, show that although FlexiEdit falls slightly short of P2P for background preservation, it surpasses all other models in CLIP Similarity.

\begin{table}[t!]
\centering
\caption{\textbf{Quantitative Comparisons in Non-rigid Editing.} We select 30 samples corresponding to non-rigid edits from the data used in the PIE benchmark \cite{proxedit} and ELITE \cite{elite} for evaluation. In Background Preservation, P2P \cite{prompt_to_prompt}, when used with Direct Inversion \cite{direct_inversion} methods, scores the highest. However, in CLIP similarity scores, FlexiEdit outperforms the other models, demonstrating superior alignment.}
\label{table:1}
\resizebox{\textwidth}{!}{
\begin{tabular}{@{} l c c S[table-format=2.2] S[table-format=2.2] S[table-format=2.2] S[table-format=2.2] S[table-format=2.2] S[table-format=2.2] @{}}
\toprule
& \multicolumn{2}{c}{Editing Flexibility} & \multicolumn{4}{c}{Background Preservation} & \multicolumn{2}{c}{CLIP Similarity} \\
\cmidrule(lr){2-3} \cmidrule(lr){4-7} \cmidrule(lr){8-9}
\textbf{Method} & \scriptsize \textbf{Rigid-edit} & \scriptsize \textbf{Non-rigid edit} & \scriptsize{$\textbf{PSNR}$ $\uparrow$} & \scriptsize{$\textbf{LPIPS}_{\times10^3}$ $\downarrow$} &{$\textbf{MSE}_{\times10^4}$ $\downarrow$} & {$\textbf{SSIM}_{\times10^2}$ $\uparrow$} & {$\textbf{Whole}$ $\uparrow$} & {$\textbf{Edited}$ $\uparrow$} \\
\midrule
PnP \cite{plug_and_play} & \checkmark & $\times$ & 22.46 & 106.06 & 80.45 & 79.68 & 22.31 & 20.62 \\
P2P \cite{prompt_to_prompt} & \checkmark & $\times$ & \bfseries 27.22 & \bfseries 54.55 & \bfseries 32.86 & \bfseries 84.76 & 22.56 & 21.10 \\
MasaCtrl \cite{masactrl} & \checkmark & \checkmark & 22.64 & 87.94 & 81.09 & 81.33 & 23.54 & 22.32 \\
ProxMasaCtrl \cite{proxedit} & \checkmark & \checkmark & 24.43 & 85.32 & 75.09 & 82.53 & 23.85 & 22.53 \\
FlexiEdit (Ours) & \checkmark & \checkmark & 25.74 & 80.45 & 58.45 & 82.62 & \bfseries 25.28 & \bfseries 24.88 \\
\bottomrule
\end{tabular}}
\end{table}

\begin{table}[ht!]
\centering
\caption{\textbf{Quantitative Comparisons in Rigid Editing.} Evaluated using the PIE benchmark \cite{proxedit}, P2P \cite{prompt_to_prompt} shows superior performance in Background Preservation, whereas FlexiEdit has the highest performance in CLIP similarity within edited regions.}
\label{table:2}
\resizebox{\textwidth}{!}{%
\begin{tabular}{@{} l S S S S S S S @{}}
\toprule
& \multicolumn{1}{c}{Structure} & \multicolumn{4}{c}{Background Preservation} & \multicolumn{2}{c}{CLIP Similarity} \\
\cmidrule(lr){2-2} \cmidrule(lr){3-6} \cmidrule(lr){7-8}
\scriptsize \textbf{Method} & \scriptsize{$\textbf{Distance}_{\times10^3}$ $\downarrow$} & \scriptsize{$\textbf{PSNR}$ $\uparrow$} & \scriptsize{$\textbf{LPIPS}_{\times10^3}$ $\downarrow$} & \scriptsize{$\textbf{MSE}_{\times10^4}$ $\downarrow$} & \scriptsize{$\textbf{SSIM}_{\times10^2}$ $\uparrow$} & \scriptsize{$\textbf{Whole}$ $\uparrow$} & \scriptsize{$\textbf{Edited}$ $\uparrow$} \\
\midrule
PnP \cite{plug_and_play} & 24.29 & 22.46 & 106.06 & 80.45 & 79.68 & \bfseries 25.41 & 22.62 \\
P2P \cite{prompt_to_prompt} & \bfseries 11.65 & \bfseries 27.22 & \bfseries 54.55 & \bfseries 32.86 & \bfseries 84.76 & 25.02 & 22.10 \\
MasaCtrl \cite{masactrl} & 24.70 & 22.64 & 87.94 &  81.09 & 81.33 & 24.38 & 21.35 \\
ProxMasaCtrl \cite{proxedit} & 22.85 & 24.43 & 85.32 & 75.09 & 82.53 & 24.75 & 21.58 \\
FlexiEdit (Ours) & 22.13 & 25.74 & 80.45 & 58.45 & 82.62 & 25.15 & \bfseries 22.87 \\
\bottomrule
\end{tabular}%
}
\end{table}

\paragraph{\bf Rigid Editing Results}
The results of rigid editing are presented in Fig \ref{fig:7_experiment_rigid_edit}, showing examples of object changes and style transfers. P2P and MasaCtrl are heavily influenced by the original layout, struggling to make significant, natural alterations. In contrast, FlexiEdit shows greater flexibility in adapting the layout, producing more natural outcomes. In Fig \ref{fig:7_experiment_rigid_edit}, FlexiEdit's results show lions, dolls, and zebras being less affected by the original objects. The qualitative results are in Table \ref{table:2}. While P2P, using the Direct Inversion \cite{direct_inversion} method, scored highest in Structure Distance and Background Preservation, FlexiEdit achieved the highest CLIP similarity for the edited region. Although FlexiEdit's background preservation falls short compared to other methods, it excels in modifying images according to user requirements.

\begin{figure}[b!]
\centering
    \includegraphics[width=0.9\textwidth]{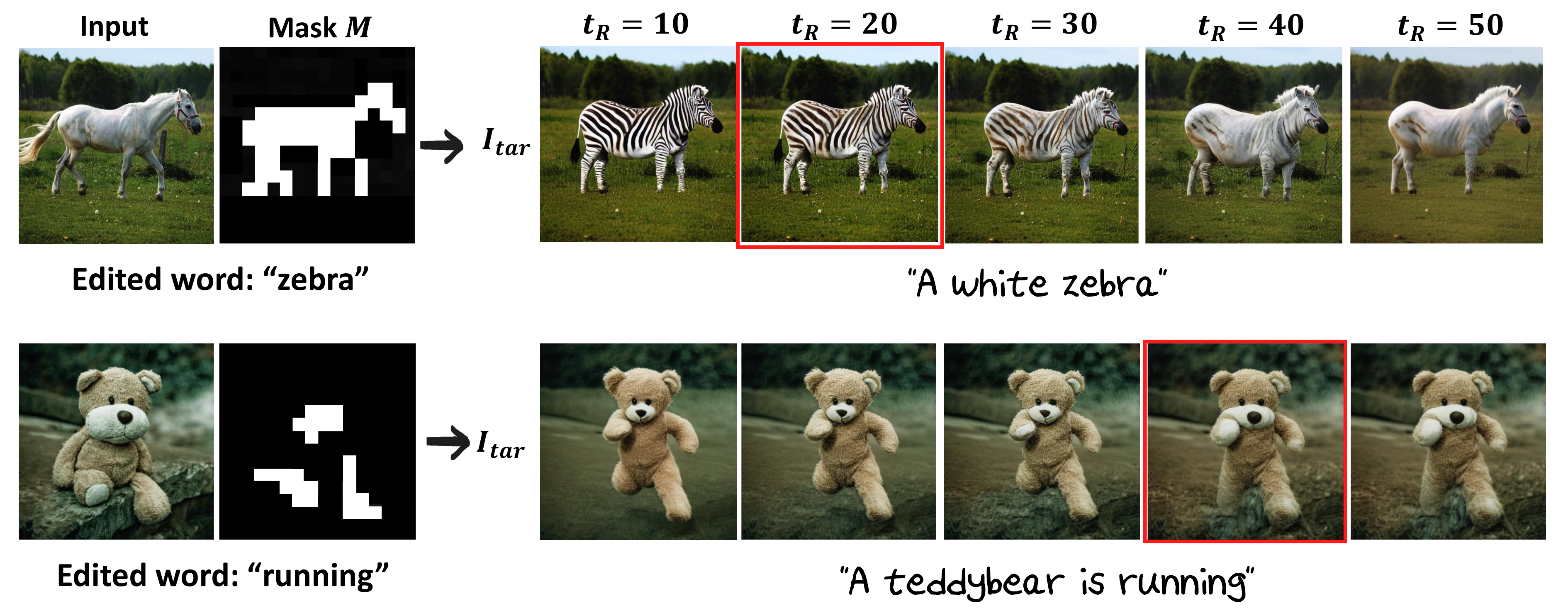}
   \caption{\textbf{Image outcomes by re-inversion duration, $t_R$}. For object changes (rigid edits), a smaller $t_R$ preserves the edited zebra's features. In contrast, for pose changes (non-rigid edits), a larger $t_R$ maintains the original image's characteristics. Thus, we set the range of $t_R$ according to the type of editing.}
\label{fig:8_ablation_t_R}
\end{figure}

\subsection{Ablation Study}
\paragraph{\bf Optimal $t_R$ Configuration for Diverse Editing Tasks}
\label{ablation:t_R}

Extensive experimentation investigates how editing outcomes vary with different $t_R$ values during the Re-inversion process. Our findings indicate that the optimal range of $t_R$ values depends on the type of editing. Generally, a shorter $t_R$ results in fewer attributes of the original image, whereas a longer $t_R$ preserves more of these attributes. As shown in Fig \ref{fig:8_ablation_t_R}, the first example depicts an object change from ``white horse'' to ``white zebra''. Here, a smaller $t_R$ retains more of the zebra's characteristics, while a larger $t_R$ incorporates features of the original white horse. For such object changes, setting $t_R$ within [10, 30] and applying Eq \ref{eq:13} yields the best result at $t_R=20$. Conversely, in the second example, where a teddy bear is edited to be ``running'', the object and background took longer to assimilate features from the original image. For these non-rigid edits, setting $t_R$ within [30, 50] and applying Eq \ref{eq:13} found that an optimal $t_R$ is 38. In conclusion, the range for $t_R$ varies with the editing type. Object changes within rigid edits benefit from setting $t_R$ within [10, 30], while edits requiring maximal preservation of the original image, as well as non-rigid edits, show improved results with $t_R$ set within [30, 50].

\begin{figure}[t!]
\centering
    \includegraphics[width=0.9\textwidth]{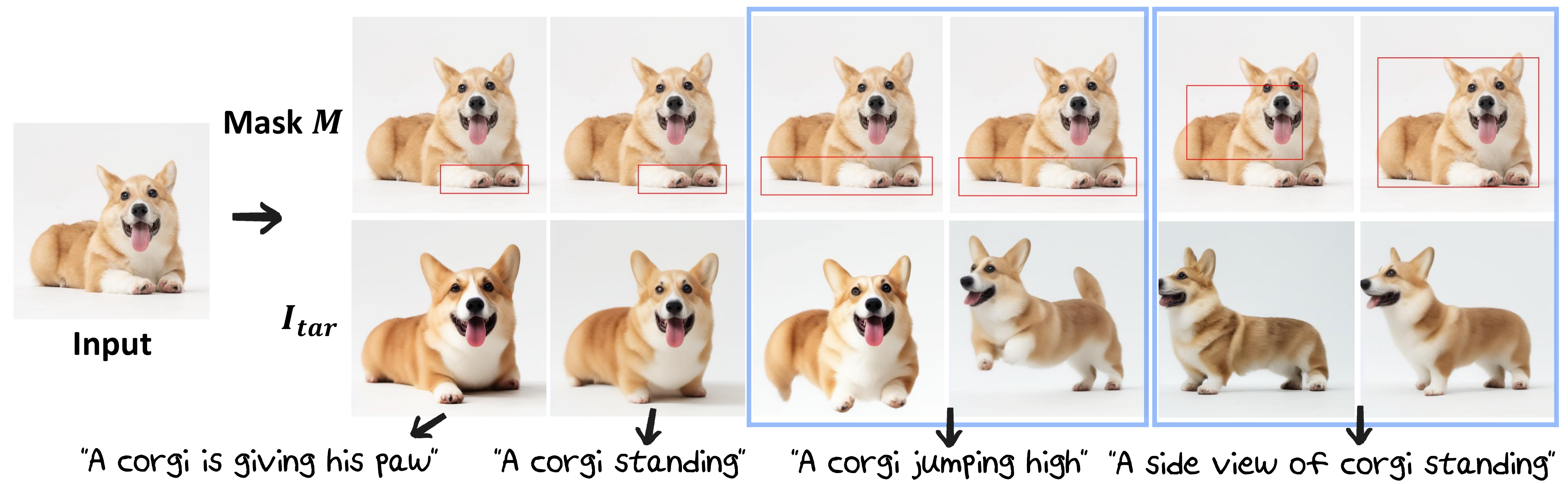}
   \caption{\textbf{Image outcomes based on the size of user mask $M$ (red box)}. When $M$ is small, changes occur within $M$ while preserving the original layout. As $M$ enlarges, adding Gaussian noise to the DDIM latent intensifies, resulting in a new layout.}
\label{fig:8_ablation_mask}
\end{figure}

\paragraph{\bf Impact of User-Defined Mask Regions for Image Editing}
While deriving mask $M$ from edited words through cross-attention is convenient, it can be challenging to refine latent features in user-desired locations. Hence, providing a user mask is advantageous for precise image edits. We conducted an ablation study to explore how images can be edited with various user-defined masks. The results in Fig \ref{fig:8_ablation_mask} show that when the mask region is narrow, areas outside the mask are preserved. Enlarging the mask region progressively departs from the original image's layout. As the mask enlarges, the area for latent refinement expands, making the edit increasingly independent of the original layout.

\section{Conclusion}

In this paper, we propose FlexiEdit, a method that allows for more flexible editing of the original image's layout. FlexiEdit achieves this by reducing high-frequency components in the DDIM latent, enabling a wider range of edits, and utilizing a three-branch scheme to better reflect the characteristics of the original image. Compared to other image editing methods, FlexiEdit demonstrates superior performance in non-rigid editing and offers more flexible layout changes in rigid editing, aligning better with the user's input text prompt. We believe that FlexiEdit addresses the shortcomings of existing image editing methods and contributes to a more advanced and versatile editing framework.


\clearpage  

\section*{Acknowledgements}
This work was partly supported by Institute for Information \& communications Technology Planning \& Evaluation (IITP) grant funded by the Korea government(MSIT) (No. 2021-0-01381, Development of Causal AI through Video Understanding and Reinforcement Learning, and Its Applications to Real Environments) and partly supported by Institute of Information \& communications Technology Planning \& Evaluation (IITP) grant funded by the Korea government(MSIT) (No.2022-0-00184, Development and Study of AI Technologies to Inexpensively Conform to Evolving Policy on Ethics).

%
%
\bibliographystyle{splncs04}
\bibliography{main}

\begin{thebibliography}{10}
\providecommand{\url}[1]{\texttt{#1}}
\providecommand{\urlprefix}{URL }
\providecommand{\doi}[1]{https://doi.org/#1}

\bibitem{blended_diffusion}
Avrahami, O., Lischinski, D., Fried, O.: Blended diffusion for text-driven editing of natural images. In: Proceedings of the IEEE/CVF Conference on Computer Vision and Pattern Recognition. pp. 18208--18218 (2022)

\bibitem{gan_2019}
Brock, A., Donahue, J., Simonyan, K.: Large scale gan training for high fidelity natural image synthesis (2019)

\bibitem{masactrl}
Cao, M., Wang, X., Qi, Z., Shan, Y., Qie, X., Zheng, Y.: Masactrl: Tuning-free mutual self-attention control for consistent image synthesis and editing. arXiv preprint arXiv:2304.08465  (2023)

\bibitem{diffedit}
Couairon, G., Verbeek, J., Schwenk, H., Cord, M.: Diffedit: Diffusion-based semantic image editing with mask guidance. arXiv preprint arXiv:2210.11427  (2022)

\bibitem{gan_overview}
Creswell, A., White, T., Dumoulin, V., Arulkumaran, K., Sengupta, B., Bharath, A.A.: Generative adversarial networks: An overview. IEEE signal processing magazine  \textbf{35}(1),  53--65 (2018)

\bibitem{diffusion_beat_gans}
Dhariwal, P., Nichol, A.: Diffusion models beat gans on image synthesis. Advances in neural information processing systems  \textbf{34},  8780--8794 (2021)

\bibitem{tokenflow}
Geyer, M., Bar-Tal, O., Bagon, S., Dekel, T.: Tokenflow: Consistent diffusion features for consistent video editing. arXiv preprint arXiv:2307.10373  (2023)

\bibitem{gan_2014}
Goodfellow, I.J., Pouget-Abadie, J., Mirza, M., Xu, B., Warde-Farley, D., Ozair, S., Courville, A., Bengio, Y.: Generative adversarial networks (2014)

\bibitem{proxedit}
Han, L., Wen, S., Chen, Q., Zhang, Z., Song, K., Ren, M., Gao, R., Stathopoulos, A., He, X., Chen, Y., et~al.: Proxedit: Improving tuning-free real image editing with proximal guidance. In: Proceedings of the IEEE/CVF Winter Conference on Applications of Computer Vision. pp. 4291--4301 (2024)

\bibitem{prompt_to_prompt}
Hertz, A., Mokady, R., Tenenbaum, J., Aberman, K., Pritch, Y., Cohen-Or, D.: Prompt-to-prompt image editing with cross attention control. arXiv preprint arXiv:2208.01626  (2022)

\bibitem{ddpm}
Ho, J., Jain, A., Abbeel, P.: Denoising diffusion probabilistic models. Advances in neural information processing systems  \textbf{33},  6840--6851 (2020)

\bibitem{cfg}
Ho, J., Salimans, T.: Classifier-free diffusion guidance. arXiv preprint arXiv:2207.12598  (2022)

\bibitem{direct_inversion}
Ju, X., Zeng, A., Bian, Y., Liu, S., Xu, Q.: Direct inversion: Boosting diffusion-based editing with 3 lines of code. arXiv preprint arXiv:2310.01506  (2023)

\bibitem{imagic}
Kawar, B., Zada, S., Lang, O., Tov, O., Chang, H., Dekel, T., Mosseri, I., Irani, M.: Imagic: Text-based real image editing with diffusion models. In: Proceedings of the IEEE/CVF Conference on Computer Vision and Pattern Recognition. pp. 6007--6017 (2023)

\bibitem{diffusion_clip}
Kim, G., Kwon, T., Ye, J.C.: Diffusionclip: Text-guided diffusion models for robust image manipulation. In: Proceedings of the IEEE/CVF Conference on Computer Vision and Pattern Recognition. pp. 2426--2435 (2022)

\bibitem{wavelet}
Koo, G., Yoon, S., Yoo, C.D.: Wavelet-guided acceleration of text inversion in diffusion-based image editing. In: ICASSP 2024-2024 IEEE International Conference on Acoustics, Speech and Signal Processing (ICASSP). pp. 4380--4384. IEEE (2024)

\bibitem{stylediffusion}
Li, S., van~de Weijer, J., Hu, T., Khan, F.S., Hou, Q., Wang, Y., Yang, J.: Stylediffusion: Prompt-embedding inversion for text-based editing. arXiv preprint arXiv:2303.15649  (2023)

\bibitem{npi}
Miyake, D., Iohara, A., Saito, Y., Tanaka, T.: Negative-prompt inversion: Fast image inversion for editing with text-guided diffusion models. arXiv preprint arXiv:2305.16807  (2023)

\bibitem{nti}
Mokady, R., Hertz, A., Aberman, K., Pritch, Y., Cohen-Or, D.: Null-text inversion for editing real images using guided diffusion models. In: Proceedings of the IEEE/CVF Conference on Computer Vision and Pattern Recognition. pp. 6038--6047 (2023)

\bibitem{glide}
Nichol, A., Dhariwal, P., Ramesh, A., Shyam, P., Mishkin, P., McGrew, B., Sutskever, I., Chen, M.: Glide: Towards photorealistic image generation and editing with text-guided diffusion models. arXiv preprint arXiv:2112.10741  (2021)

\bibitem{fatezero}
Qi, C., Cun, X., Zhang, Y., Lei, C., Wang, X., Shan, Y., Chen, Q.: Fatezero: Fusing attentions for zero-shot text-based video editing. arXiv preprint arXiv:2303.09535  (2023)

\bibitem{clip}
Radford, A., Kim, J.W., Hallacy, C., Ramesh, A., Goh, G., Agarwal, S., Sastry, G., Askell, A., Mishkin, P., Clark, J., et~al.: Learning transferable visual models from natural language supervision. In: International conference on machine learning. pp. 8748--8763. PMLR (2021)

\bibitem{dalle2}
Ramesh, A., Dhariwal, P., Nichol, A., Chu, C., Chen, M.: Hierarchical text-conditional image generation with clip latents (2022)

\bibitem{ldm}
Rombach, R., Blattmann, A., Lorenz, D., Esser, P., Ommer, B.: High-resolution image synthesis with latent diffusion models. In: Proceedings of the IEEE/CVF conference on computer vision and pattern recognition. pp. 10684--10695 (2022)

\bibitem{imagen}
Saharia, C., Chan, W., Saxena, S., Li, L., Whang, J., Denton, E., Ghasemipour, S.K.S., Ayan, B.K., Mahdavi, S.S., Lopes, R.G., Salimans, T., Ho, J., Fleet, D.J., Norouzi, M.: Photorealistic text-to-image diffusion models with deep language understanding (2022)

\bibitem{ddim}
Song, J., Meng, C., Ermon, S.: Denoising diffusion implicit models. arXiv preprint arXiv:2010.02502  (2020)

\bibitem{structure_distance}
Tumanyan, N., Bar-Tal, O., Bagon, S., Dekel, T.: Splicing vit features for semantic appearance transfer. In: Proceedings of the IEEE/CVF Conference on Computer Vision and Pattern Recognition. pp. 10748--10757 (2022)

\bibitem{plug_and_play}
Tumanyan, N., Geyer, M., Bagon, S., Dekel, T.: Plug-and-play diffusion features for text-driven image-to-image translation. In: Proceedings of the IEEE/CVF Conference on Computer Vision and Pattern Recognition. pp. 1921--1930 (2023)

\bibitem{ssim}
Wang, Z., Bovik, A.C., Sheikh, H.R., Simoncelli, E.P.: Image quality assessment: from error visibility to structural similarity. IEEE transactions on image processing  \textbf{13}(4),  600--612 (2004)

\bibitem{elite}
Wei, Y., Zhang, Y., Ji, Z., Bai, J., Zhang, L., Zuo, W.: Elite: Encoding visual concepts into textual embeddings for customized text-to-image generation. arXiv preprint arXiv:2302.13848  (2023)

\bibitem{tune_a_video}
Wu, J.Z., Ge, Y., Wang, X., Lei, S.W., Gu, Y., Shi, Y., Hsu, W., Shan, Y., Qie, X., Shou, M.Z.: Tune-a-video: One-shot tuning of image diffusion models for text-to-video generation. In: Proceedings of the IEEE/CVF International Conference on Computer Vision. pp. 7623--7633 (2023)

\bibitem{freeinit}
Wu, T., Si, C., Jiang, Y., Huang, Z., Liu, Z.: Freeinit: Bridging initialization gap in video diffusion models. arXiv preprint arXiv:2312.07537  (2023)

\bibitem{object_aware_inversion}
Yang, Z., Gui, D., Wang, W., Chen, H., Zhuang, B., Shen, C.: Object-aware inversion and reassembly for image editing. arXiv preprint arXiv:2310.12149  (2023)

\bibitem{frag}
Yoon, S., Koo, G., Kim, G., Yoo, C.D.: Frag: Frequency adapting group for diffusion video editing. arXiv preprint arXiv:2406.06044  (2024)

\bibitem{stackgan}
Zhang, H., Xu, T., Li, H., Zhang, S., Wang, X., Huang, X., Metaxas, D.N.: Stackgan: Text to photo-realistic image synthesis with stacked generative adversarial networks. In: Proceedings of the IEEE international conference on computer vision. pp. 5907--5915 (2017)

\bibitem{psnr_lpips}
Zhang, R., Isola, P., Efros, A.A., Shechtman, E., Wang, O.: The unreasonable effectiveness of deep features as a perceptual metric. In: Proceedings of the IEEE conference on computer vision and pattern recognition. pp. 586--595 (2018)

\end{thebibliography}

\newpage

\appendix
\begin{center}
    \Large \textbf{Supplementary Material for FlexiEdit:} \\
    \textbf{Frequency-Aware Latent Refinement for \\ Enhanced Non-Rigid Editing}
\end{center}

\section{Implementation Details}
\subsection{Selection of Edited Words}
In this section, we aim to provide a detailed explanation of the process for selecting edited words, as briefly mentioned in Section 4.1. Consider an input image $I_{src}$ with source prompt $p_{src}$ and a target prompt $p_{tar}$, aiming for an edit towards $I_{tar}$. The words from $p_{src}$ that are intended to be modified on $I_{src}$ are represented as $w_{ed}$. The initial phase of the selection process involves comparing $p_{src}$ and $p_{tar}$ to identify common terms. The removal of these overlapping terms yields an intermediate set, termed as $w'_{ed}$.  

\begin{align}
& w'_{ed} = p_{src} \setminus (p_{src} \cap p_{tar}), \label{eq_ablation:1}
\end{align}


In Eq \ref{eq_ablation:1}, set notation is employed, with the set difference $\setminus$ indicating words unique to $p_{src}$ compared with $p_{tar}$, and $\cap$ identifying common words between the two prompts. Let’s represent the similarity between an image and text using the CLIP similarity \cite{clip} score, symbolized as $\text{CLIP}_{sim}$. Setting $\text{CLIP}_{sim}(I_{src}, p_{src})$ as the threshold, when $w'_{ed}$ consists of N words, we calculate $\text{CLIP}_{sim}(I_{src}, w'_{ed})$ for each. Instances of $w'_{ed}$ that yield a similarity score lower than the threshold are designated as $w_{ed}$, representing the edited words. This selection process can be expressed as follows:

\begin{align}
& w_{ed} = \left\{ w'_{ed_i} \mid \text{CLIP}_{sim}(I, w'_{ed_i}) < \text{CLIP}_{sim}(I, p_{src}), \, i = 1, 2, ..., N \right\}, \label{eq_ablation:2}
\end{align}

\subsection{Mask Extraction}
During the DDIM inversion process with $p_{tar}$ and $I_{src}$ as inputs, we compute cross-attention maps at a resolution of $16 \times 16$ from every layer of the UNet. Given the inversion progresses through timesteps $t=1$ to $T$, and with $N_{p_{tar}}$ indicating the number of words in $p_{tar}$, the cross-attention maps across the timesteps can be represented as follows:


\begin{align}
& {[c^{{p_{tar}}}_t]}^T_{t=1} \in \mathbb{R}^{16 \times 16 \times N_{p_{tar}} \times T}, \label{eq_ablation:3}
\end{align}

In our experiments, the cross-attention map generated at t=1, which reflects the image before any noise introduction, is effectively used to identify areas significantly relevant to the words in $p_{tar}$. When $w_{ed}$ comprises $N_w$ words, we derive an average cross-attention map, denoted as $\bar{c}^{w_{ed}} \in \mathbb{R}^{16 \times 16 \times 1}$, by averaging the $N_w$ maps from the $c^{p_{tar}}_{t=1}$ that corresponds to $w_{ed}$ (Eq. \ref{eq_ablation:4}). This average map $\bar{c}^{w_{ed}}$ initially a $16 \times 16$ grayscale image, is normalized to scale the values between 0 and 1 (Eq. \ref{eq_ablation:5}). Subsequently, applying a threshold of 0.3, we transform it into a binary image (Eq. \ref{eq_ablation:6}). This binary image is upsampled to $64 \times 64$ to produce the final mask, $M$, effectively highlighting areas of interest for the editing process based on the edited words (Eq. \ref{eq_ablation:7}).

\begin{align}
&\bar{c}^{w_{ed}} = \frac{1}{N_w} \sum_{i=1}^{N_w} c^{w_{ed_i}}_{t=1} \in \mathbb{R}^{16 \times 16 \times 1}, \label{eq_ablation:4} \\
& \bar{c}_{norm} = \text{Normalize}(\bar{c}^{w_{ed}}, 0, 1), \label{eq_ablation:5} \\
& \bar{c}_{binary} = 
\begin{cases} 
1 & \text{if } \bar{c}_{norm} > 0.3, \\
0 & \text{otherwise}.
\end{cases}, \label{eq_ablation:6} \\
& M = \text{Upsample}(\bar{c}_{binary}, 64 \times 64), \label{eq_ablation:7} 
\end{align}

\subsection{Analysis of High Frequency Reduction Impact}


\begin{figure}[t!]
\centering
    \includegraphics[width=\textwidth]{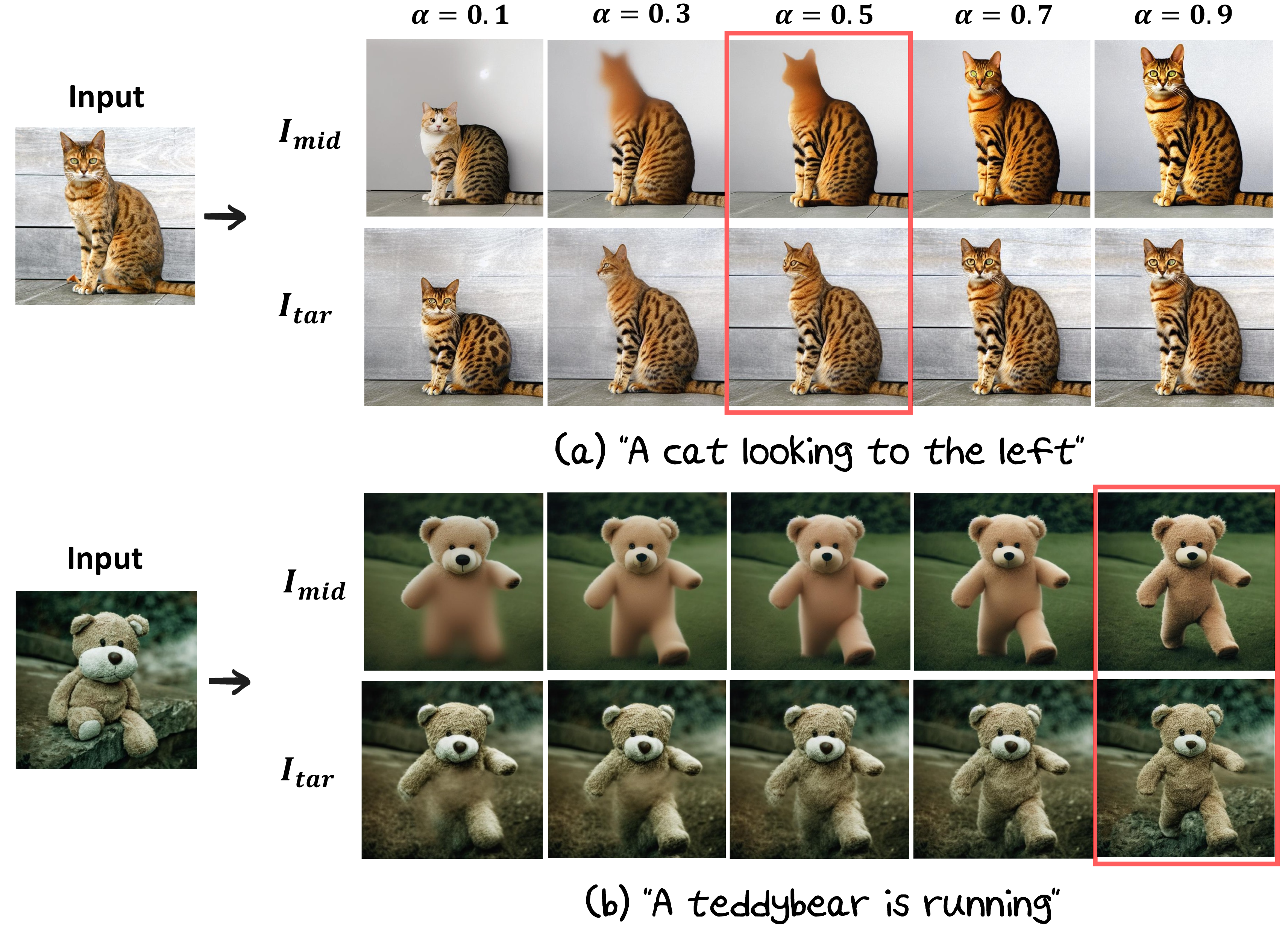}
   \caption{\textbf{Ablation Study on $\alpha$ Values.} (a) The edited region forms around the cat's face. (b) The edited region appears around the teddy bear's legs. The size of the edited region increases from (a) to (b), with the optimal alpha values reflecting this progression: $0.5$ in (a) and $0.9$ in (b), demonstrating a similar tendency of increase.}
\label{fig_ab:1_alpha}
\end{figure}


Expanding on the discussion of $\alpha$ values from Section 4.1, we present adjustments to $\alpha$ across diverse images in Fig \ref{fig_ab:1_alpha}. This figure demonstrates the progression of the edited region's size from (a) to (b), with the optimal $\alpha$ values being $0.5$ for (a) and $0.9$ for (b). Specifically, for the smaller edited region in Fig \ref{fig_ab:1_alpha} (a), setting $\alpha$ below $0.5$ leads to notable changes in the original image, which can result in the loss of background details. Conversely, settings $\alpha$ above $0.9$ tend to preserve the original image's layout and characteristics more effectively. In the case of the larger edited region in Fig \ref{fig_ab:1_alpha} (b), we observe that setting $\alpha$ below $0.5$ can cause blurring in the teddy bear's legs. In contrast, an $\alpha$ value of 0.9 prevents blurring, maintaining clarity in the edited region. This underscores the importance of adjusting $\alpha$ in accordance with the size of the edited region to achieve optimal editing results. This led us to discover a tendency where the optimal $\alpha$ value decreases with smaller edited regions and increases as the edited region enlarges. Therefore, setting the range of $\alpha$ as $[\alpha_{min}, \alpha_{max}]$ and denoting the total area of mask $M$ as $A_{total}$ and the area marked as $1$ within mask $M$ (the region requiring editing) as $A_{edit}$, we calculate $\alpha$ using the Eq \ref{eq_ablation:8}. Based on the results from analyzing a diverse set of samples, $\alpha_{min}$ is set to $0.5$ and $\alpha_{max}$ to $0.9$. Specifically, this formula ensures that once the ratio $A_{edit}/A_{total}$ exceeds 0.5, $\alpha$ is set to the $\alpha_{max}$ value.


\begin{align}
&\alpha = 
\begin{cases} 
\alpha_{\text{min}} + (2 \times (\alpha_{\text{max}} - \alpha_{\text{min}}) \times \left(\frac{A_{\text{edit}}}{A_{\text{total}}}\right)) & \text{if } \frac{A_{\text{edit}}}{A_{\text{total}}} \leq 0.5, \\
\alpha_{\text{max}} & \text{if } \frac{A_{\text{edit}}}{A_{\text{total}}} > 0.5 , \label{eq_ablation:8} 
\end{cases}
\end{align}

We observe that when the edited region is small, reducing high-frequency components within a limited area can make it challenging to alter the layout of objects significantly. To facilitate noticeable changes in the layout within these smaller regions, it's essential to reduce high-frequency components substantially. Conversely, in larger edited regions, a slight reduction in high-frequency components can more easily modify the original layout of objects. Therefore, the size of the edited region directly influences the optimal $\alpha$ value, with smaller regions necessitating lower $\alpha$ values and larger regions benefiting from higher $\alpha$ values to yield desirable outcomes.


\section{More Qulitative Comparison}
\begin{figure}[b!]
\begin{center}
\centerline{\includegraphics[width=0.95\textwidth]{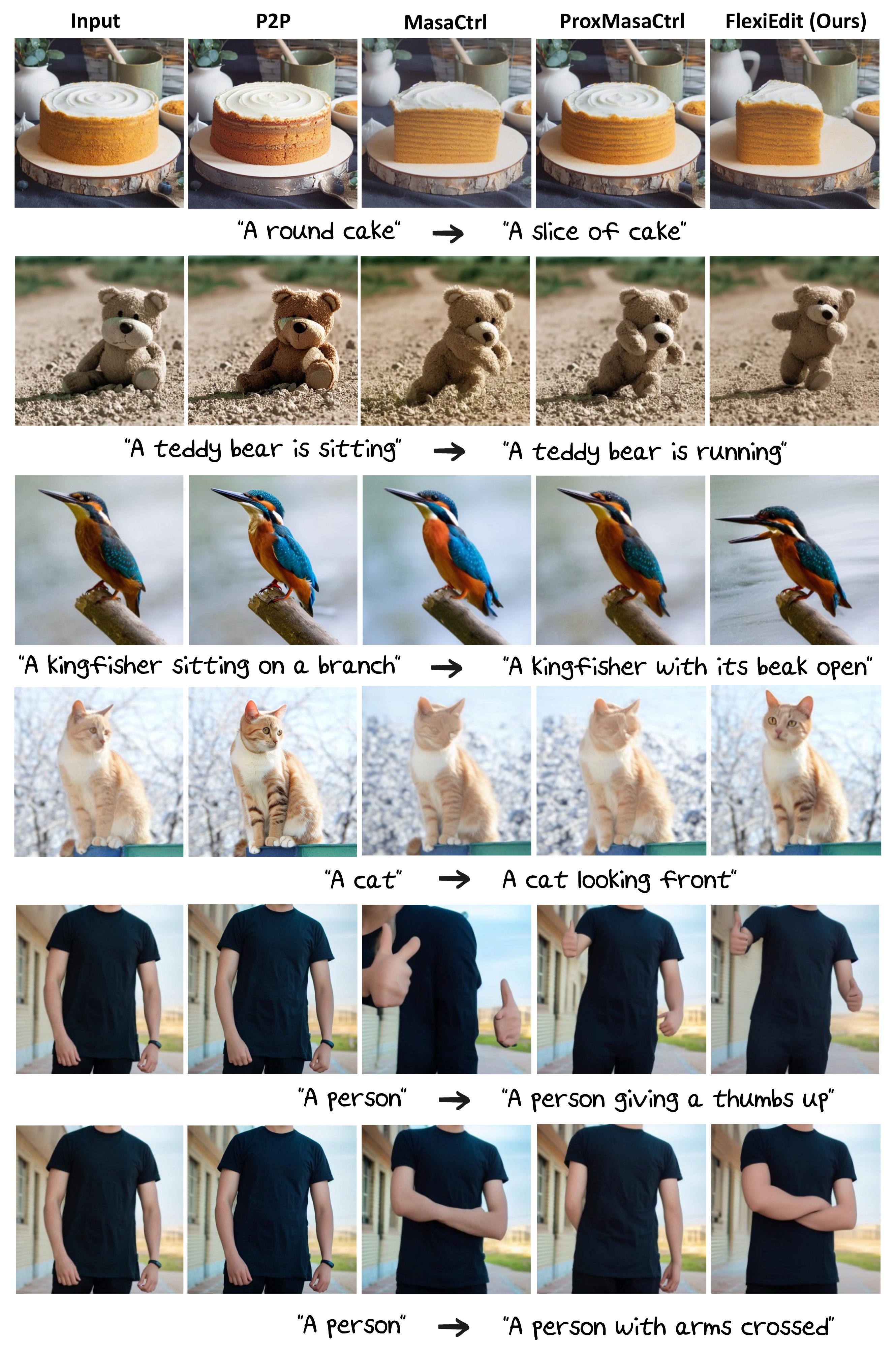}}
\caption{ \textbf{Additional Qualitative Comparison in Non-Rigid Editing}. Demonstrates FlexiEdit's superior performance in non-rigid editing tasks over current image editing methods such as P2P \cite{prompt_to_prompt}, MasaCtrl \cite{masactrl}, and ProxMasaCtrl \cite{proxedit}
}
\label{fig_ab:3_non_rigid_edit}
\end{center}
\end{figure}

\begin{figure}[b!]
\begin{center}
\centerline{\includegraphics[width=0.95\textwidth]{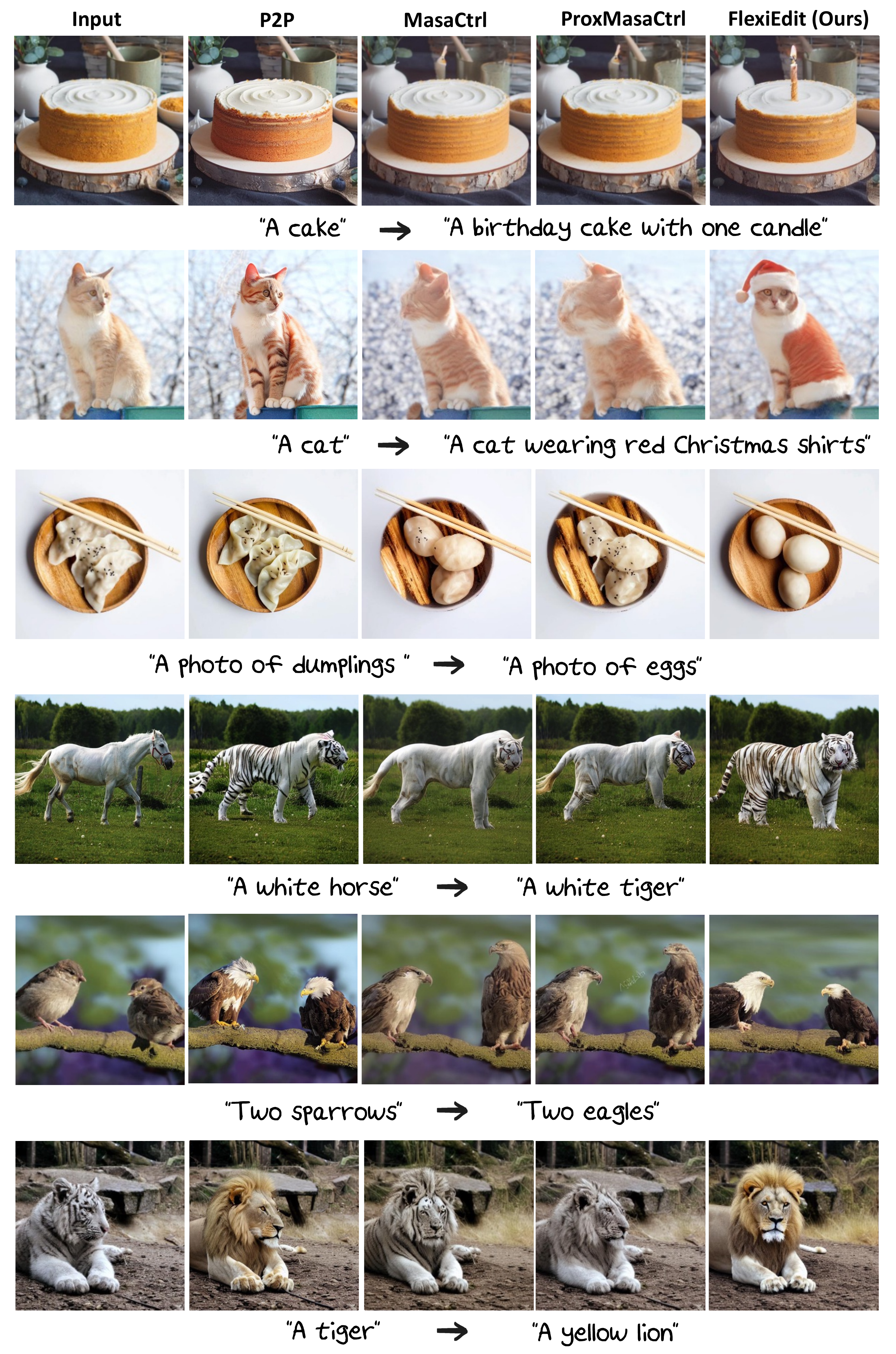}}
\caption{ \textbf{Additional Qualitative Comparison in Rigid Editing.} Illustrates how FlexiEdit can more flexibly alter the original image layout, delivering results in rigid editing that align more closely with the text input compared to other methods.
}
\label{fig_ab:4_rigid_edit}
\end{center}
\end{figure}

Additional results for non-rigid editing are presented in Fig \ref{fig_ab:3_non_rigid_edit}, while outcomes for rigid editing are shown in Fig \ref{fig_ab:4_rigid_edit}. Comparative models include P2P \cite{prompt_to_prompt}, MasaCtrl \cite{masactrl}, ProxMasaCtrl \cite{proxedit}, and FlexiEdit. Notably, P2P employs the NTI \cite{nti} inversion method. FlexiEdit stands out in both non-rigid and rigid edits, significantly surpassing existing models by flexibly transforming the original image layout and closely matching the user's textual input in the edits.

\section{Limitations and Future Works}

\begin{figure}[t!]
\centerline{\includegraphics[width=\textwidth]{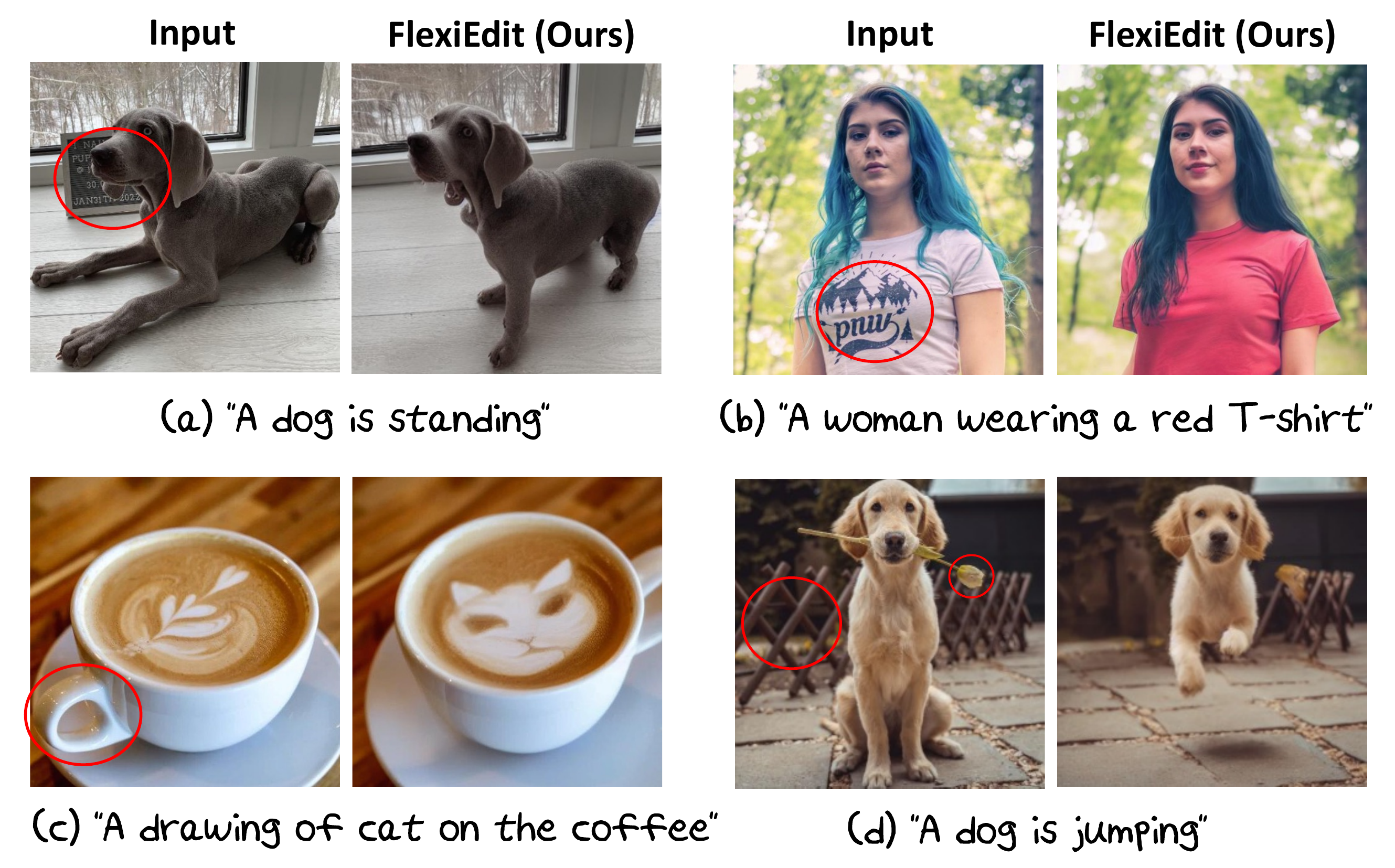}}
\caption{ \textbf{FlexiEdit Failure Cases.} Results showcasing loss of background and detail in the original images. (a) An object behind the dog disappears, (b) text on the woman's shirt is not preserved, (c) the cup's handle is reversed, and (d) a fence behind the dog and the flower held by the dog are not retained.
}
\label{fig_ab:2_failure_case}
\end{figure}

FlexiEdit possesses the advantage of reducing high-frequency components in the edit area, allowing for more flexible alterations of the original image layout. However, there have been instances where the background or details of the original image are not perfectly preserved. Such examples can be observed in Fig \ref{fig_ab:2_failure_case}, wherein (a) an object behind the dog disappeared, (b) patterns on the woman's dress are lost, (c) the direction of the cup's handle changed, and (d) despite the dog jumping, background details are omitted.

As mentioned in Section $2.2$, this issue emerges because when the CFG scale exceeds $1$ during the DDIM sampling process, it deviates from the original DDIM Inversion trajectory. Moreover, as FlexiEdit utilizes a refined DDIM latent $z^{'}_{T}$, it diverges further from the original DDIM latent $z_T$. While using inversion methods like NTI \cite{nti} or Direct Inversion \cite{direct_inversion} to force the DDIM sampling trajectory to align with the DDIM Inversion trajectory can preserve the background and details of the original image, it restricts the flexibility of edits, such as altering the layout of objects. We conclude there is a trade-off between fidelity, which aims to preserve the original image, and editability, focused on enabling flexible changes to the layout. Therefore, future research on FlexiEdit should focus on expanding in a direction that allows for flexible layout changes while ensuring regions outside the edited area maintain high fidelity with the original image.

\section{Evaluation Metrics}
In the evaluation of FlexiEdit compared to other editing methods, we leverage six metrics applied to images from PIE-bench\cite{proxedit} and ELITE\cite{elite}. These metrics are selected to provide a comprehensive assessment of editing quality, focusing on structural integrity, background preservation, visual fidelity, and textual consistency. Below is a detailed explanation of each metric used in our evaluation:

\textbf{Structure Distance\cite{structure_distance}}: This metric is designed to assess the structural integrity between the original and edited images by analyzing the self-similarity of deep spatial features, specifically extracted from DINO-ViT models. By measuring the cosine similarity of these features, the structure distance focuses on the preservation of the image's structural essence rather than its aesthetic elements. Such an approach is particularly effective for evaluating image editing tasks, which aim to maintain the core structural composition without inducing significant alterations.

\textbf{PSNR, LPIPS \cite{psnr_lpips}}: These metrics assess the quality of background preservation in the edited images. Peak Signal-to-Noise Ratio (PSNR) measures the pixel-level accuracy, providing a quantitative evaluation of noise introduced through editing. Learned Perceptual Image Patch Similarity (LPIPS) offers insights into perceptual similarity, evaluating how perceptually close the edited image is to the original, thus accounting for human visual perception nuances.

\textbf{MSE, SSIM \cite{ssim}}: Mean Squared Error (MSE) and Structural Similarity Index Measure (SSIM) are utilized to assess the fidelity and visual quality of the edits. MSE quantifies the average squared difference between the edited and original images, serving as a direct measure of error magnitude. SSIM evaluates changes in texture, contrast, and structure, providing a measure of how these visual elements are preserved or altered through the editing process.

\textbf{CLIP similarity \cite{clip}}: To ensure the edited images remain consistent with the textual prompts, CLIP similarity is employed. This metric measures the semantic alignment between the text descriptions and the visual content of the edited images. It ensures that the edits are contextually relevant and aligned with the intended modifications, enhancing the edit's overall coherence and relevance. Evaluations are conducted on both the entire image and the edited regions specifically, offering a detailed analysis of text-image consistency.

\clearpage
\newpage

\clearpage
\newpage

\end{document}